\documentclass[letterpaper,10pt,conference]{ieeeconf}

\usepackage[T1]{fontenc}
\usepackage[utf8]{inputenc}
\usepackage{array}
\usepackage{makecell}
\usepackage{graphicx}

\newif\ifoptimizedfigures
\optimizedfigurestrue
\ifoptimizedfigures
  \graphicspath{{src/figure_optimized/}{src/figure/}}
  \DeclareGraphicsExtensions{.jpg,.jpeg,.png,.pdf}
\else
  \graphicspath{{src/figure/}}
  \DeclareGraphicsExtensions{.png,.pdf,.jpg,.jpeg}
\fi
\usepackage{amsmath,amssymb}
\usepackage{soul}
\usepackage{booktabs}
\usepackage{tikz}
\usepackage{pgf-pie}
\usepackage{pgfplots}
\pgfplotsset{compat=1.18}
\usepackage{rotating}
\usepackage{subcaption}
\usepackage{url}
\usepackage{cite}
\usepackage{titlesec}
\titlespacing*{\section}{0pt}{*1.5}{*0.8}
\titlespacing*{\subsection}{0pt}{*1.2}{*0.6}
\usepackage{tabularx}
\usepackage{multirow}
\usepackage{xcolor}
\usepackage{colortbl}
\usepackage{pifont}
\usepackage{threeparttable}
\usepackage{hyperref}
\usepackage{xurl}  
\usepackage{indentfirst}
\usepackage{dblfloatfix} 

\usepackage{aprl_misc}
\usepackage{aprl_acronyms}

\usetikzlibrary{shapes.geometric, arrows.meta, positioning, calc, fit, backgrounds}

\newcommand{\yes}{\textcolor{green!60!black}{\ding{51}}}
\newcommand{\no}{\textcolor{red!70!black}{\ding{55}}}
\newcommand{\pmark}{$\circ$}
\newcommand{\tightfloatvspace}{\vspace{-0.7em}}

\IEEEoverridecommandlockouts

\title{\LARGE \bf
MarsLab: A Martian Rover Simulator for \\ Planetary Rover Autonomous Navigation
}

\author{
Hoyun Kim$^{1*}$, Beomsu Kim$^{1}$, and Giseop Kim$^{1\dagger}$%
\thanks{$^{1}$H. Kim, B. Kim, and G. Kim are with the Department of Robotics and
Mechatronics Engineering, DGIST, Daegu, Republic of Korea
{\ttfamily\small \{hoyunkim, beomsu.kim, gsk\}@dgist.ac.kr}.}%
\thanks{$^{*}$First author: Hoyun Kim. $^{\dagger}$Corresponding author: Giseop Kim.}%
\thanks{This work was supported by the National Research Foundation of Korea (NRF) grants funded by the Korea government (MSIT) (No. RS-2026-25492530 and No. RS-2026-25517444), and by the InnoCORE program of the Ministry of Science and ICT (No. 26-InnoCORE-01).}%
}

\begin{document}
\bstctlcite{IEEEtranBSTCTL:etal3}

\maketitle
\thispagestyle{empty}
\pagestyle{empty}

\begin{abstract}
Future Mars missions will require rover autonomy that can operate across
unstructured terrain, changing illumination, atmospheric dust, and limited
communication.  Simulation is a practical way to study these conditions before
deployment, but existing Mars-relevant resources differ in scope, including
mission-oriented simulators, fixed analog datasets, task-specific environments,
and open robotics interfaces.  In this context, we present MarsLab, an open-source, ROS2-native
Mars rover simulator for autonomy and navigation algorithm development.  MarsLab
combines HiRISE-derived and procedural terrain with customizable rock, crater,
solar-illumination, and atmospheric-dust settings, and runs a
Perseverance-class rover model in NVIDIA Isaac Sim.  The runtime publishes RGB,
depth, RGB-D point clouds, LiDAR, IMU, wheel odometry, and \ac{GT} pose data
through standard ROS2 topics.  We demonstrate MarsLab with \ac{SLAM} benchmarks
across sensing modalities, dust levels, scene geometry, and route length, and
with \ac{VPR} benchmarks over repeated Mars Base traversals under illumination
and dust changes.  The results illustrate how controlled scene variation and
shared \ac{GT} trajectories can be used to compare trajectory-level estimation
and image-level place recognition within the same simulator.  Our Project Page:
\url{https://kimhoyun-robotair.github.io/MarsLab/}.
\end{abstract}

\definecolor{figoneaccent}{HTML}{55B85A}
\definecolor{figonedark}{HTML}{1F2424}
\definecolor{figonemid}{HTML}{5B6262}
\definecolor{figoneline}{HTML}{D2D8D5}
\definecolor{figonepanel}{HTML}{F7F8F6}

\begin{figure}[t]
  \centering
  \resizebox{\columnwidth}{!}{%
  \begin{tikzpicture}[
      x=1cm,y=1cm,
      font=\sffamily,
      panel/.style={rounded corners=2pt, draw=figoneline, line width=0.45pt, fill=figonepanel},
      img/.style={inner sep=0pt, outer sep=0pt},
      tag/.style={rounded corners=1.2pt, fill=black!72, text=white,
                  inner xsep=5pt, inner ysep=2.5pt, font=\scriptsize\bfseries},
      note/.style={font=\small, text=figonedark, align=left},
      herotag/.style={rounded corners=1.2pt, fill=black!72, text=white,
                     inner xsep=6pt, inner ysep=3pt, font=\small\bfseries},
    ]

    \def\W{8.80}
    \def\HeroW{8.48}
    \def\TileW{2.72}
    \def\TileH{2.16}
    \def\AlgW{4.12}
    \def\AlgH{1.92}

    \node[panel, minimum width=\W cm, minimum height=10.50cm, anchor=north west] (outer) at (0,0) {};

    \node[note, font=\small\bfseries, anchor=west] at (0.18,-0.28)
      {What We Aim For};

    \node[img, anchor=north west] (hero) at (0.16,-0.55)
      {\includegraphics[width=\HeroW cm]{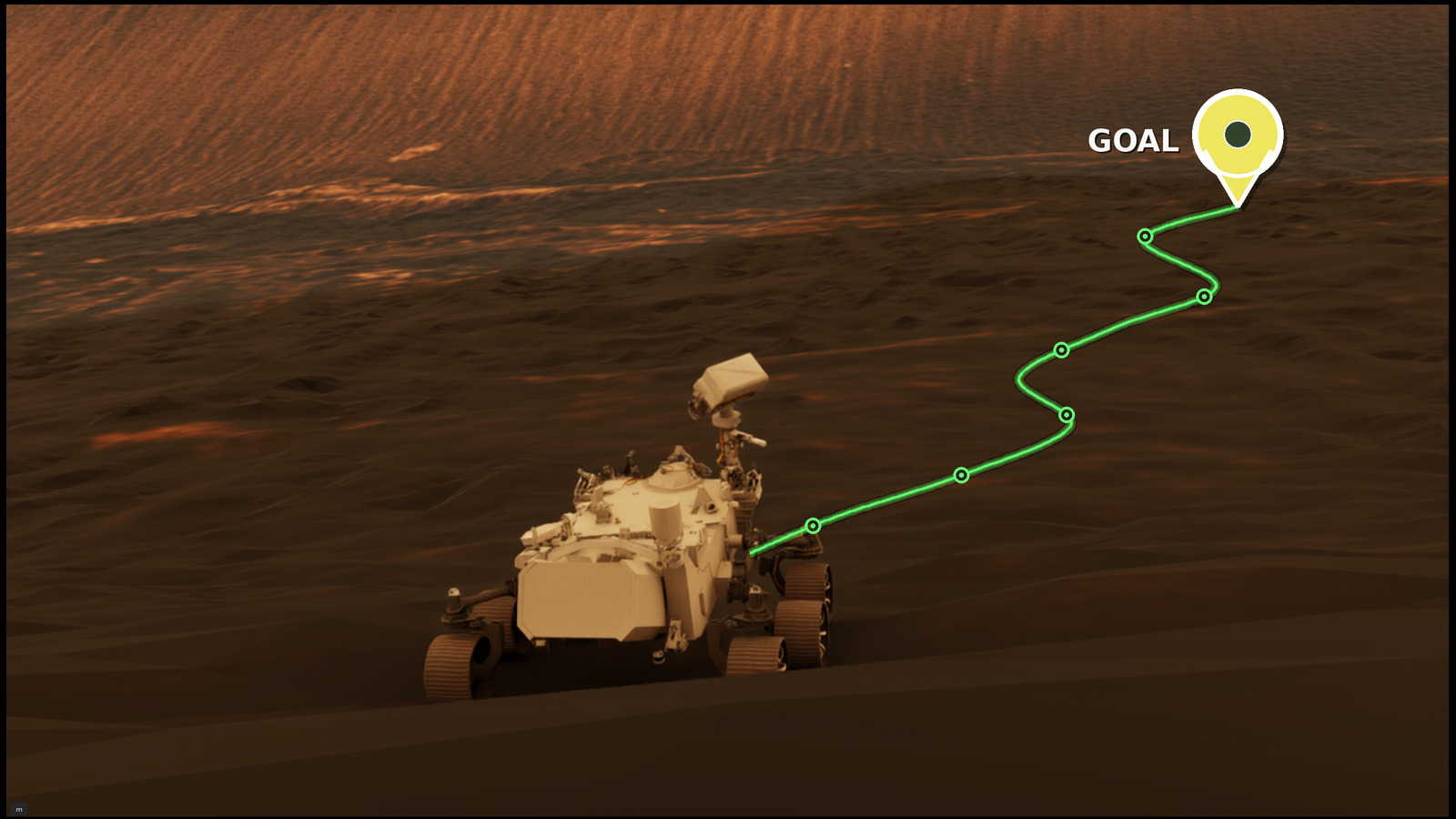}};
    \draw[figonedark, line width=0.35pt] (hero.north west) rectangle (hero.south east);
    \node[herotag, anchor=north west] at (hero.north west)
      {Navigation Task};
    \node[anchor=south east, rounded corners=1.2pt, fill=white!90, draw=figoneline,
          line width=0.35pt, inner xsep=6pt, inner ysep=3pt,
          font=\small\bfseries, text=figonedark] at (hero.south east)
      {Planned Route + Goal};

    \node[note, font=\small\bfseries, anchor=west] at (0.18,-5.62)
      {What We Built};
    \draw[figoneline, line width=0.55pt] (2.80,-5.62) -- (8.46,-5.62);

    \foreach \x/\name/\file in {
      0.16/RGB Camera/fig1_rgb,
      3.04/Depth/fig1_depth,
      5.92/3D LiDAR/fig1_lidar} {
      \node[panel, minimum width=\TileW cm, minimum height=\TileH cm, anchor=north west] at (\x,-5.78) {};
      \node[img, anchor=north west] at (\x+0.12,-5.93)
        {\includegraphics[width=2.48cm,height=2.02cm,keepaspectratio]{fig01/\file}};
      \node[tag, anchor=north west] at (\x+0.12,-5.93) {\name};
    }

    \node[note, font=\small\bfseries, anchor=west] at (0.18,-8.22)
      {What We Can Evaluate};
    \draw[figoneline, line width=0.55pt] (3.98,-8.22) -- (8.46,-8.22);

    \node[panel, minimum width=\AlgW cm, minimum height=\AlgH cm, anchor=north west] (slamPanel) at (0.16,-8.43) {};
    \node[panel, minimum width=\AlgW cm, minimum height=\AlgH cm, anchor=north west] (vprPanel) at (4.52,-8.43) {};

    \node[img, anchor=north west] (slamImg) at (0.26,-8.49)
      {\includegraphics[width=3.92cm,height=1.80cm]{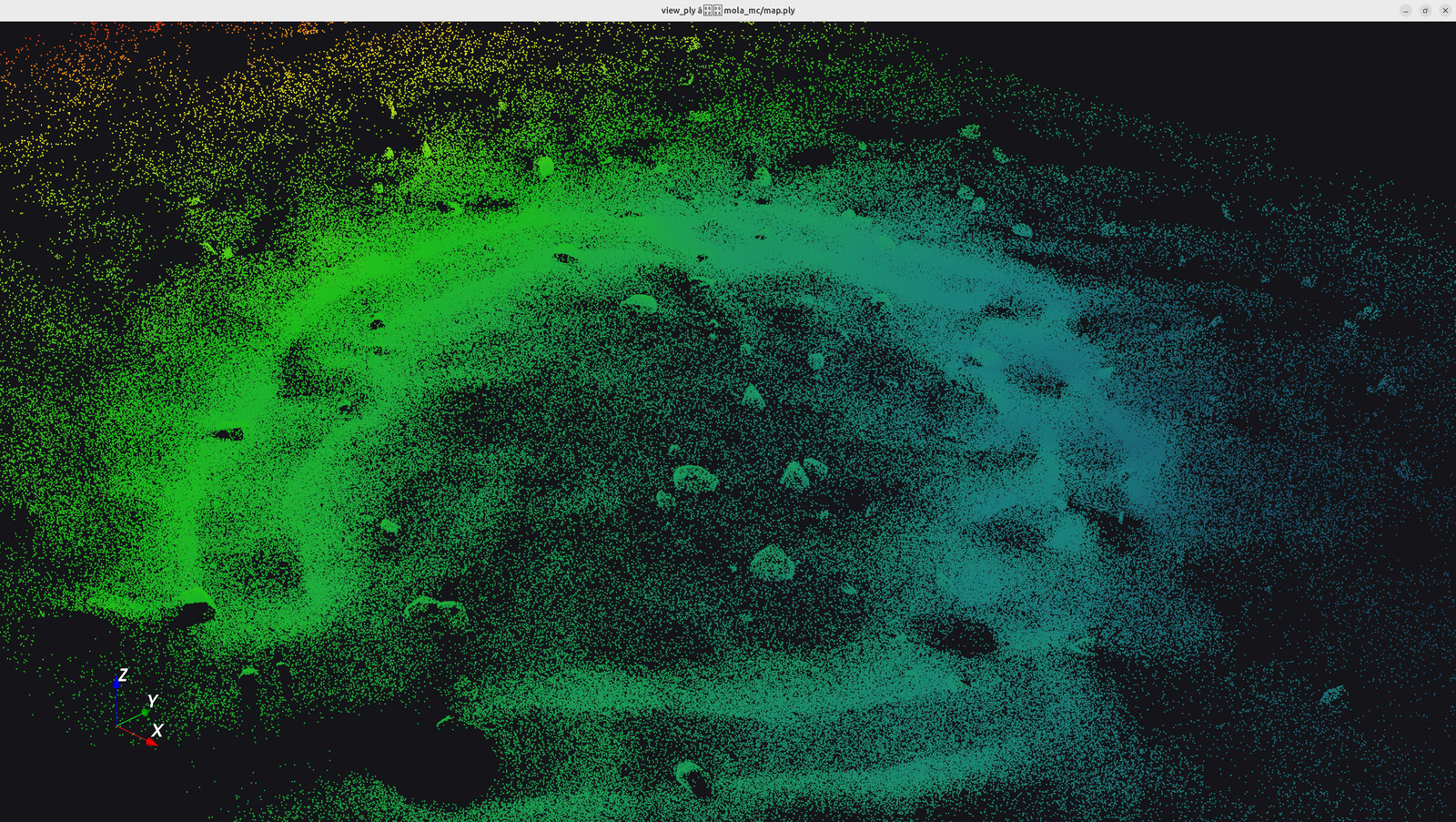}};
    \node[tag, anchor=north west] at (slamImg.north west) {SLAM};

    \node[img, anchor=north west] (vprA) at (4.62,-8.49)
      {\includegraphics[width=1.90cm,height=1.80cm]{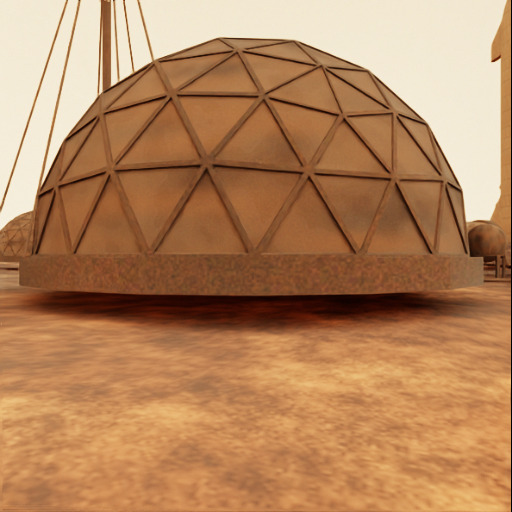}};
    \node[img, anchor=north west] (vprB) at (6.64,-8.49)
      {\includegraphics[width=1.90cm,height=1.80cm]{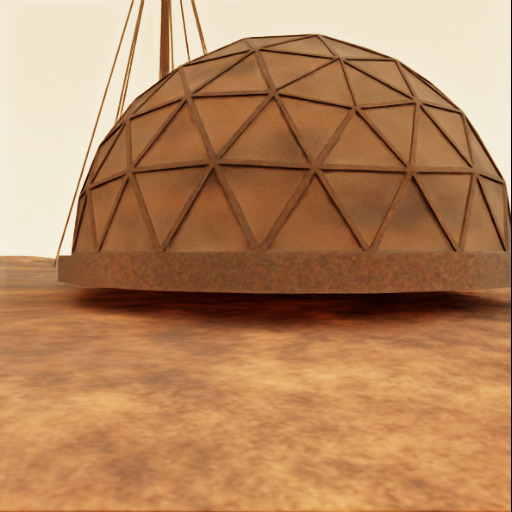}};
    \draw[figoneline, line width=0.55pt] (6.58,-8.49) -- (6.58,-10.29);
    \node[tag, anchor=north west] at (vprA.north west) {VPR};
  \end{tikzpicture}}
  \caption{\textbf{MarsLab overview.}  MarsLab supports Mars-rover autonomy and
  navigation algorithm development through HiRISE-based customizable scenes and
  environment variables, together with sensor topic streams.  Users can vary
  terrain, rocks, craters, solar illumination, and atmospheric dust
  (\figref{fig:scenes-proc}--\figref{fig:dust}), subscribe to RGB, depth/RGB-D,
  and 3D~LiDAR sensor topics from the rover runtime, and evaluate autonomy
  modules including SLAM and visual place recognition, as demonstrated in our
  experiments.}
  \label{fig:overview}
\tightfloatvspace
\end{figure}

\section{Introduction}
\label{sec:intro}


Mars exploration is entering a renewed robotic phase.  In 2026, NASA approved
implementation of its support to ESA's Rosalind Franklin rover and selected
SpaceX Falcon Heavy for the mission's late-2028 launch
opportunity~\cite{nasa2026rosa}.  NASA also announced Space Reactor-1 Freedom,
a nuclear-electric spacecraft intended to reach Mars before the end of 2028 and
deploy a Skyfall payload of Ingenuity-class
helicopters~\cite{nasa2026nationalpolicy}.  Together with Perseverance AutoNav
and Ingenuity, these plans point toward Mars missions that combine rover
mobility, aerial scouting, commercial launch, and increasingly autonomous
robotic decision making~\cite{verma2023perseverance,jpl2023futuremarshelicopter}.

Mars exploration remains difficult despite this momentum.  Mission systems have
progressed from Mars Exploration Rover visual
odometry~\cite{maimone2007twoyearsvo} to Perseverance driving
autonomy~\cite{verma2023perseverance}, orbital-map
localization~\cite{nash2024censible}, and coupled SLAM/navigation
pipelines~\cite{geromichalos2020slam}.  Yet surface autonomy must operate
without \ac{GNSS}, with delayed communication, and over low-texture regolith,
repetitive terrain, changing illumination, dust, and tight resource
budgets~\cite{gao2017spacerobotics}.  Localization is especially limiting:
pose error propagates into mapping and planning, while repeatable
ground-truth pose data on Mars remains scarce.

\begin{table*}[t]
\centering
\caption{Concise survey of Mars rover simulation platforms and benchmark support.\protect\\
Column abbreviations and symbols: \textbf{gen.} = generation; \textbf{seqs.} = sequences;
\textbf{sens.} = sensor modalities; \textbf{\yes} = reported/core capability;
\textbf{\pmark} = partial, indirect, or not primary; \textbf{\no} = not reported;
\textbf{D} = depth; \textbf{L} = 3D LiDAR; \textbf{seg.} = semantic or instance labels;
\textbf{NR} = not reported.}
\label{tab:mars-sim-survey}
\vspace{0.5mm}
\scriptsize
\setlength{\tabcolsep}{2.6pt}
\renewcommand{\arraystretch}{1}
\newcommand{\surveyrowstrut}{\rule[-12pt]{0pt}{28pt}}
\resizebox{\textwidth}{!}{%
\begin{tabular}{>{\surveyrowstrut}lclllllccccc}
\specialrule{0.8pt}{0pt}{0pt}
Simulator
& Year
& Scene gen.
& Engine
& Goal
& Scenes/Seqs.
& Rock
& Dust
& Light
& Sens.
& ROS2
& Open \\
\hline
\arrayrulecolor{black!18}

\textbf{ROAMS~\cite{roams}}
& 2004
& \makecell[l]{Mars-like terrain \&\\mission simulation}
& \makecell[l]{Custom /\\JPL DARTS-DSHELL}
& \makecell[l]{Rover mobility,\\dynamics, onboard\\SW V\&V}
& \makecell[l]{Mission/testbed\\scenarios; Monte\\Carlo possible}
& \yes
& \no
& \pmark
& \makecell[l]{Stereo cameras;\\IMU; encoders;\\sun sensors}
& \no
& \pmark \\
\hline

\textbf{ENav~\cite{toupet2020enavsim}}
& 2020
& \makecell[l]{Mars 2020 ENav\\test terrains}
& \makecell[l]{ROS-based +\\HDSim/RSVP}
& \makecell[l]{Autonomous nav\\V\&V}
& \makecell[l]{Mars 2020 test\\scenarios / Monte\\Carlo runs}
& \yes
& \no
& \pmark
& \makecell[l]{Stereo/NavCam\\point cloud;\\camera rendering}
& \no
& \no \\
\hline

\textbf{Giubilato et al.~\cite{giubilato2020planetarysim}}
& 2020
& \makecell[l]{Simulated Martian /\\planetary-like\\environment}
& \makecell[l]{ROS/Gazebo}
& \makecell[l]{Visual \& LiDAR\\SLAM evaluation}
& \makecell[l]{Long/Short simulated\\rover sequences /\\datasets}
& \yes
& \no
& \pmark
& \makecell[l]{Cam / 3D L;\\stereo + LiDAR\\seqs.}
& \no
& \yes \\
\hline

\textbf{MarsSim~\cite{jiang2022marssim}}
& 2023
& \makecell[l]{Multiscale Mars\\terrain simulation}
& ROS/Gazebo
& \makecell[l]{High-fidelity\\physical \& visual\\rover simulation}
& \makecell[l]{Pahrump Hills-like\\and generated scenes;\\released seqs. NR}
& \yes
& \pmark
& \yes
& \makecell[l]{RGB/visual;\\locomotion data;\\D/L NR}
& \no
& \no \\
\hline

\textbf{ISMRS~\cite{wan2024ismrs}}
& 2024
& \makecell[l]{Digital-twin and\\asset-based scenes}
& \makecell[l]{Isaac Sim}
& \makecell[l]{Mars rover\\simulation + ML\\data synthesis}
& \makecell[l]{Real-world\\recordings +\\synthetic data}
& \yes
& \no
& \yes
& \makecell[l]{RGB/stereo;\\2D LiDAR; IMU;\\wheel odom; seg.}
& \yes
& \no \\
\hline

\textbf{RLRoverLab~\cite{mortensen2024rlroverlab}}
& 2024
& \makecell[l]{Synthetic rover\\training scenes}
& Isaac Lab
& \makecell[l]{RL navigation /\\manipulation /\\control}
& \makecell[l]{RL task scenes}
& \pmark
& \no
& \pmark
& \makecell[l]{Task-dependent;\\RGB cam, height scan,\\Isaac obs.}
& \no
& \yes \\

\arrayrulecolor{black}
\hline
\textbf{MarsLab (Ours)}
& \textbf{2026}
& \makecell[l]{\textbf{HiRISE-based +}\\\textbf{customizable}}
& \makecell[l]{\textbf{Isaac Sim}\\\textbf{ROS2}}
& \makecell[l]{\textbf{Rover autonomy}\\\textbf{algorithm}\\\textbf{development}}
& \makecell[l]{\textbf{Customizable}\\\textbf{multi-scenes;}\\\textbf{SLAM/VPR/nav tests}}
& \textbf{\yes}
& \textbf{\yes}
& \textbf{\yes}
& \makecell[l]{\textbf{RGB/D/L;}\\\textbf{wheel odom;}\\\textbf{GT logs}}
& \textbf{\yes}
& \textbf{\yes} \\
\specialrule{1.1pt}{0pt}{0pt}

\end{tabular}}
\vspace{-7.772pt}
\tightfloatvspace
\end{table*}

Because these developments cannot be reproduced through field testing on Mars,
simulation-based validation is essential for evaluating autonomy stacks before
deployment.  However, as summarized in Table~\ref{tab:mars-sim-survey},
existing resources rarely combine openness with full-stack autonomy and
navigation support: mission simulators are typically not released as community
tools~\cite{roams,toupet2020enavsim}, several prior environments emphasize
narrower task settings such as \ac{SLAM}, photorealism, \ac{RL}, or synthetic
data~\cite{giubilato2020planetarysim,jiang2022marssim,mortensen2024rlroverlab,wan2024ismrs},
and analog datasets provide fixed recordings~\cite{lamarre2020canadian}.

To address this gap, we present \textbf{MarsLab}: an open-source, ROS2~\cite{ros2026ros}-native
Mars rover simulator for repeatable autonomy evaluation across customizable
terrain, lighting, dust, sensing, rover motion, and time-aligned ground truth pose data.
MarsLab supports \ac{SLAM} and navigation algorithm development through HiRISE-based
customizable scenes~\cite{mcewen2007hirise} and standard rover sensor topics
(\figref{fig:overview}).

Our contributions are:
\begin{enumerate}
    \item \textbf{MarsLab}, an open-source, ROS2-native Mars rover simulation
          and validation environment built on NVIDIA Isaac Sim;
    \item a scientifically grounded scene-generation pipeline that couples
          HiRISE-derived terrain models with procedural Mars-like rocks,
          craters, illumination, and atmospheric dust to produce Mars-relevant
          environmental variation;
    \item a reproducible, multi-scene benchmark for \ac{SLAM}, localization,
          place recognition, planning, and navigation, demonstrated through
          \ac{SLAM} and \ac{VPR} experiments.
\end{enumerate}


\definecolor{mlpbg}{HTML}{FFFFFF}
\definecolor{mlppanel}{HTML}{F8FAFC}
\definecolor{mlpsoft}{HTML}{EEF2F6}
\definecolor{mlpline}{HTML}{CBD5E1}
\definecolor{mlptext}{HTML}{1F2937}
\definecolor{mlpmuted}{HTML}{64748B}
\definecolor{mlpblue}{HTML}{2563A8}
\definecolor{mlpteal}{HTML}{16836F}
\definecolor{mlporange}{HTML}{C56A2D}

\begin{figure*}[!t]
  \centering
  \resizebox{\textwidth}{!}{%
  \begin{tikzpicture}[
      x=1cm,y=-1cm,
      font=\sffamily,
      colpanel/.style={rounded corners=5pt, draw=mlpline, line width=0.45pt, fill=mlppanel},
      imageframe/.style={rounded corners=2.2pt, draw=mlpline, line width=0.35pt},
      whitebox/.style={rounded corners=2.2pt, draw=mlpline, line width=0.35pt, fill=white},
      flowbox/.style={rounded corners=4pt, draw=mlpline, line width=0.35pt, fill=white, align=center,
                      font=\sffamily\fontsize{5.4}{6.4}\selectfont, text=mlptext},
      title/.style={anchor=center, align=center, text=mlptext,
                    font=\sffamily\fontsize{8.2}{9.0}\selectfont\bfseries},
      subtitle/.style={anchor=north west, align=left, text=mlpmuted,
                       font=\sffamily\fontsize{5.8}{6.7}\selectfont},
      label/.style={anchor=west, align=left, text=mlptext,
                    font=\sffamily\fontsize{5.8}{6.7}\selectfont\bfseries},
      small/.style={anchor=north west, align=left, text=mlpmuted,
                    font=\sffamily\fontsize{5.2}{6.1}\selectfont},
      chip/.style={rounded corners=3pt, draw=mlpline, line width=0.35pt, fill=white,
                   text=mlptext, align=center, font=\sffamily\fontsize{6.2}{7.2}\selectfont\bfseries},
      arrow/.style={-{Latex[length=2.2mm,width=1.4mm]}, draw=mlpblue, line width=0.55pt},
    ]

    \fill[mlpbg] (-0.18,-0.18) rectangle (16.58,6.42);

    \def\cw{3.55}
    \def\ch{6.18}
    \def\xone{0.00}
    \def\xtwo{4.28}
    \def\xthree{8.56}
    \def\xfour{12.84}

    \foreach \x in {\xone,\xtwo,\xthree,\xfour}{
      \draw[colpanel] (\x,0) rectangle ++(\cw,\ch);
      \draw[mlpline, line width=0.3pt] (\x+0.18,0.86) -- (\x+\cw-0.18,0.86);
    }

    \node[title, text width=3.12cm] at (\xone+1.775,0.46) {MarsLab Assets};
    \node[title, text width=3.12cm] at (\xtwo+1.775,0.46) {Customizable\\Mars Scenes};
    \node[title, text width=3.12cm] at (\xthree+1.775,0.46) {MarsLab Runtime};
    \node[title, text width=3.12cm] at (\xfour+1.775,0.46) {Evaluation\\\& Validation};

    \begin{scope}
      \clip[rounded corners=2.2pt] (\xone+0.24,0.98) rectangle ++(3.07,1.20);
      \node[anchor=center, inner sep=0pt] at (\xone+1.775,1.58)
        {\includegraphics[width=3.58cm]{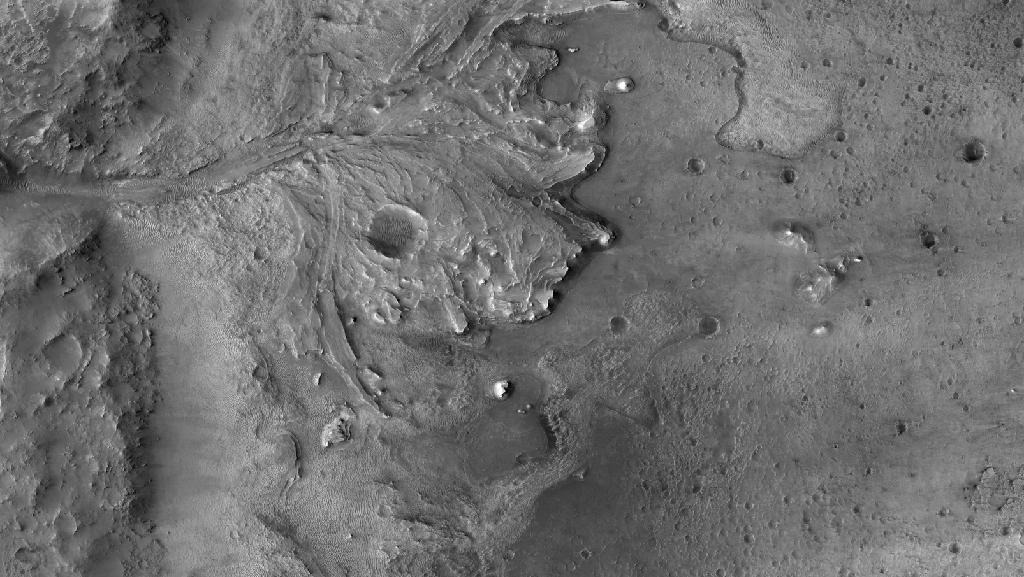}};
    \end{scope}
    \draw[imageframe] (\xone+0.24,0.98) rectangle ++(3.07,1.20);
    \node[label, anchor=center, align=center, text width=3.07cm] at (\xone+1.775,2.35) {HiRISE Terrain Base~\cite{mcewen2007hirise}};

    \begin{scope}
      \clip[rounded corners=2.2pt] (\xone+0.24,2.50) rectangle ++(1.48,1.03);
      \node[anchor=center, inner sep=0pt] at (\xone+0.98,3.015)
        {\includegraphics[width=1.66cm,height=1.03cm,keepaspectratio]{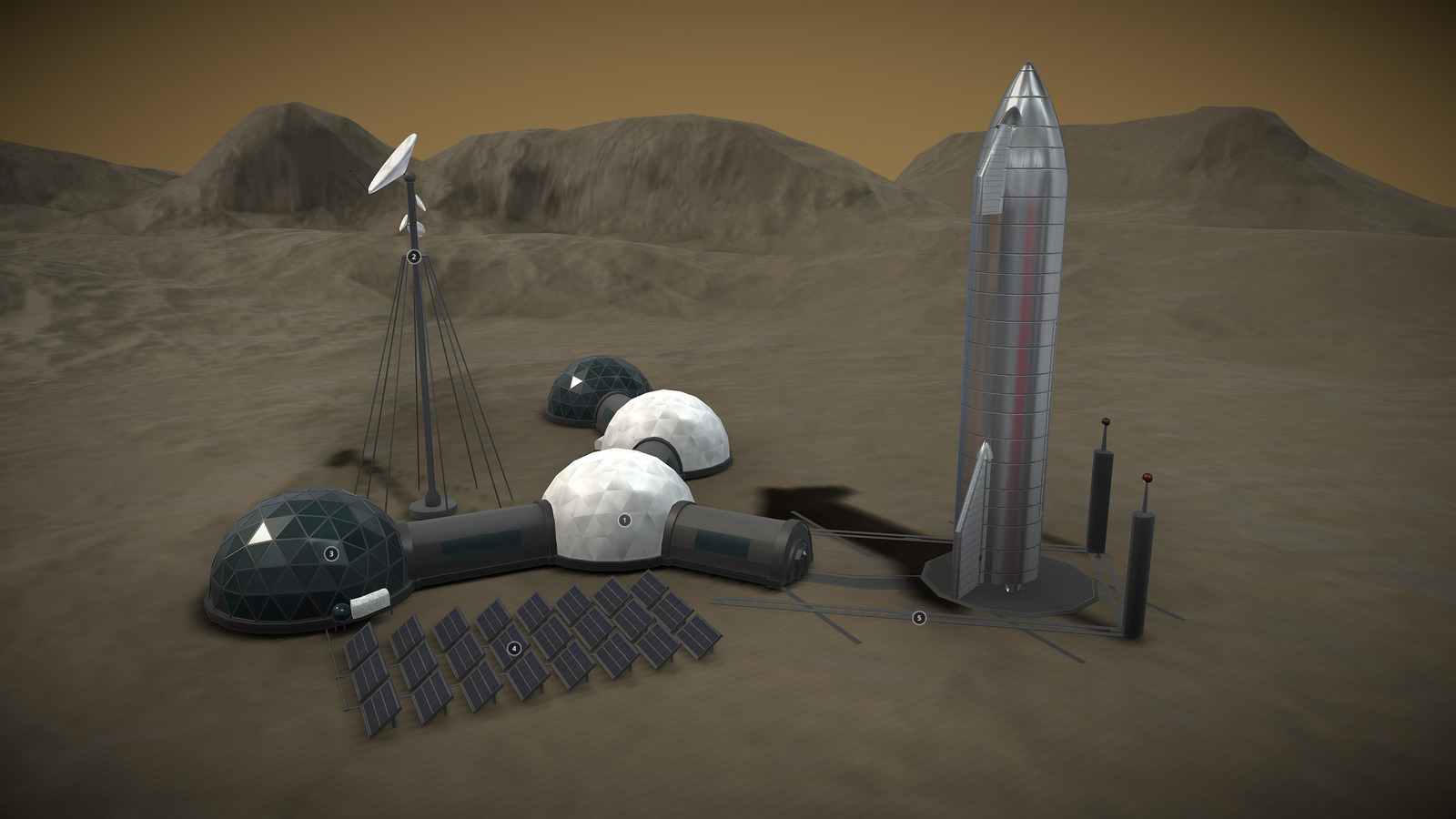}};
    \end{scope}
    \draw[imageframe] (\xone+0.24,2.50) rectangle ++(1.48,1.03);
    \begin{scope}
      \clip[rounded corners=2.2pt] (\xone+1.83,2.50) rectangle ++(1.48,1.03);
      \node[anchor=center, inner sep=0pt] at (\xone+2.57,3.015)
        {\includegraphics[width=1.66cm,height=1.03cm,keepaspectratio]{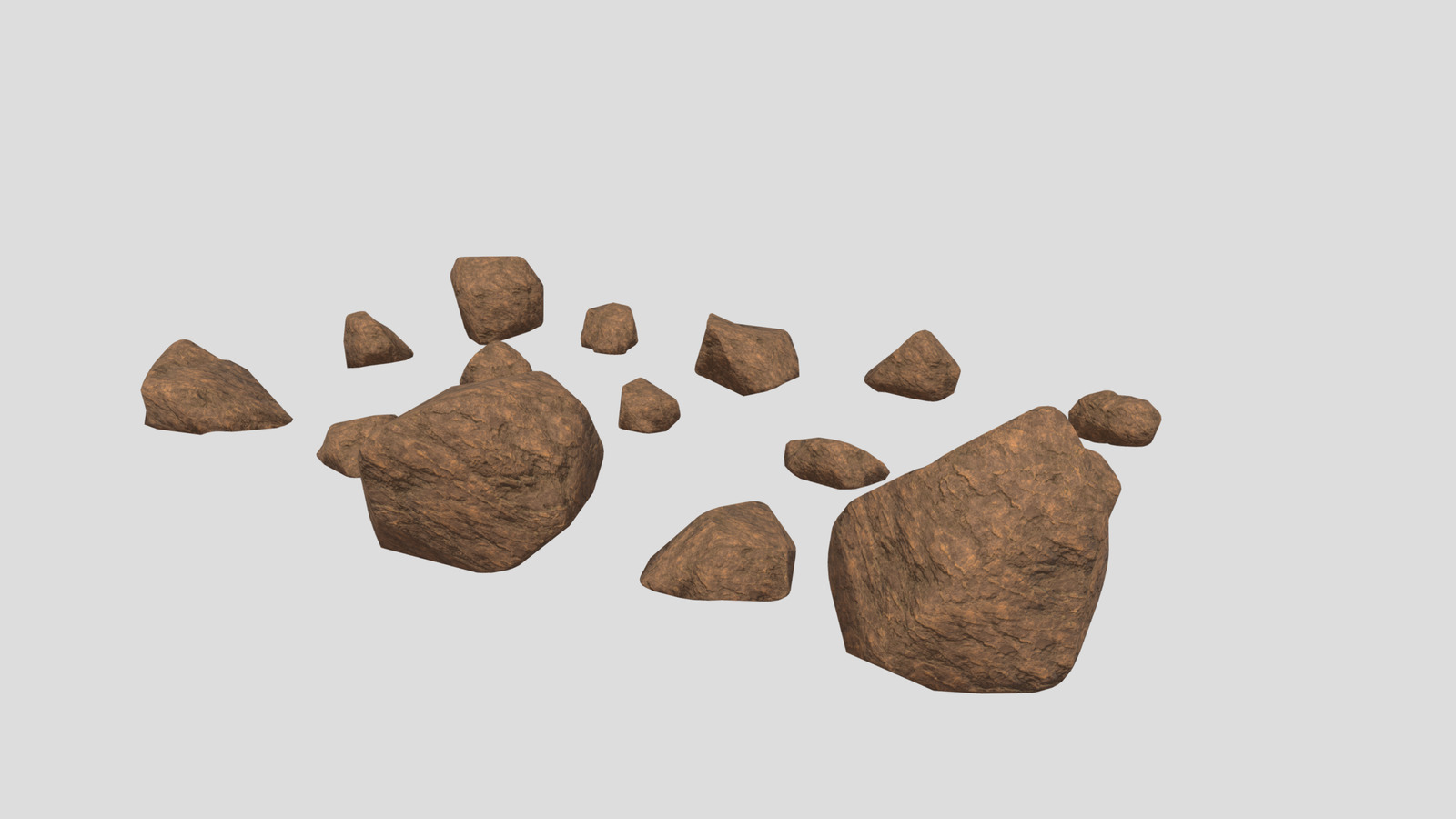}};
    \end{scope}
    \draw[imageframe] (\xone+1.83,2.50) rectangle ++(1.48,1.03);
    \node[label, anchor=center, align=center, text width=3.07cm] at (\xone+1.775,3.76) {Mars Surface Assets\hyperref[fn:mars-base]{\textsuperscript{\ref*{fn:mars-base}}}\hyperref[fn:mars-rocks]{\textsuperscript{,\ref*{fn:mars-rocks}}}};

    \begin{scope}
      \clip[rounded corners=2.2pt] (\xone+0.42,3.88) rectangle ++(2.72,1.72);
      \node[anchor=center, inner sep=0pt] at (\xone+1.775,4.74)
        {\includegraphics[width=2.72cm,height=1.72cm,keepaspectratio]{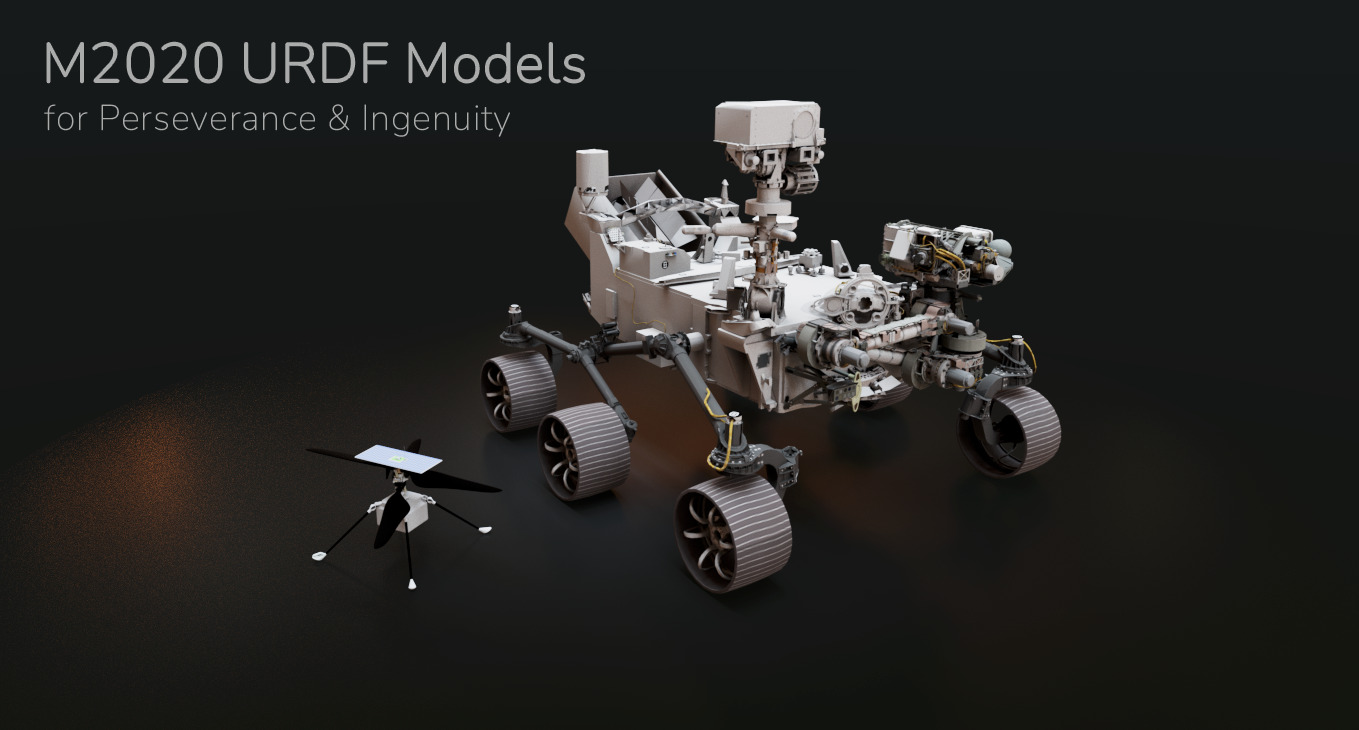}};
    \end{scope}
    \node[label, anchor=center, align=center, text width=3.07cm] at (\xone+1.775,5.70) {Rover Asset~\cite{nasajpl2022m2020urdf}};

    \foreach \i/\xx/\yy in {
      1/0.24/1.10, 2/1.85/1.10,
      3/0.24/2.16, 4/1.85/2.16,
      5/0.24/3.22, 6/1.85/3.22}{%
      \begin{scope}
        \clip[rounded corners=2.2pt] (\xtwo+\xx,\yy) rectangle ++(1.46,0.82);
        \node[anchor=center, inner sep=0pt] at (\xtwo+\xx+0.73,\yy+0.41)
          {\includegraphics[width=1.46cm,height=0.82cm,keepaspectratio]{fig02/fig2_column2_\i}};
      \end{scope}
      \draw[imageframe] (\xtwo+\xx,\yy) rectangle ++(1.46,0.82);
    }
    \node[chip, text width=2.62cm, minimum height=0.88cm, align=left,
          font=\sffamily\fontsize{5.2}{6.0}\selectfont\bfseries] at (\xtwo+1.78,4.66)
      {\textbullet\ Terrain Generation\\\textbullet\ Rock Distribution\\\textbullet\ Crater Distribution};
    \node[chip, text width=2.62cm, minimum height=0.88cm, align=left,
          font=\sffamily\fontsize{5.2}{6.0}\selectfont\bfseries] at (\xtwo+1.78,5.58)
      {\textbullet\ Illumination Change\\\textbullet\ Dust\\\textbullet\ Rover Dynamics};

    \draw[whitebox] (\xthree+0.42,1.02) rectangle ++(2.71,0.56);
    \node[anchor=center, inner sep=0pt] at (\xthree+1.08,1.30)
      {\includegraphics[width=0.44cm,height=0.44cm]{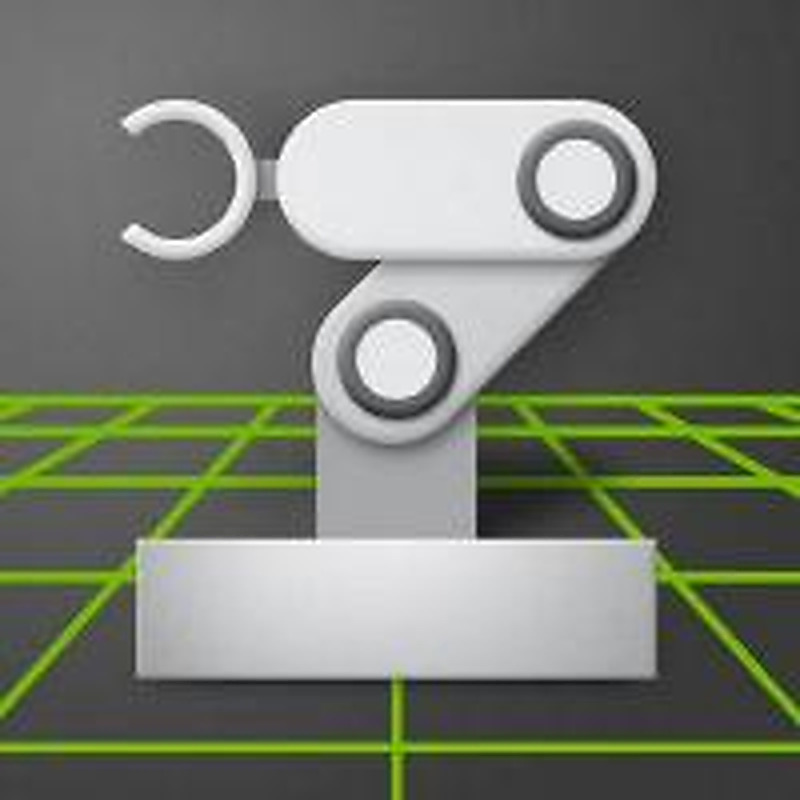}};
    \node[anchor=center, inner sep=0pt] at (\xthree+2.25,1.30)
      {\includegraphics[width=0.98cm]{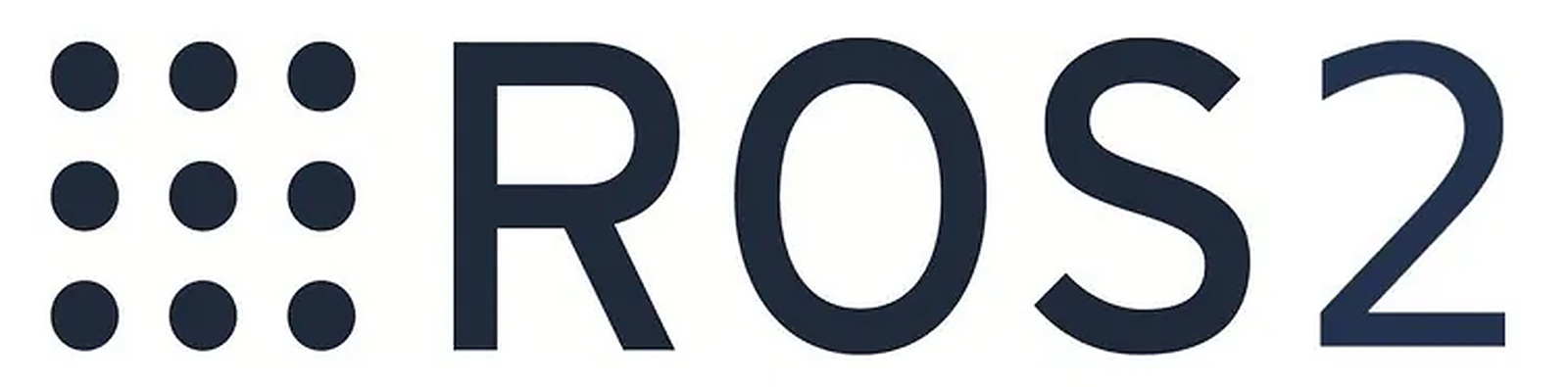}};
    \begin{scope}
      \clip[rounded corners=2.2pt] (\xthree+0.36,1.76) rectangle ++(2.83,2.00);
      \node[anchor=center, inner sep=0pt] at (\xthree+1.775,2.76)
        {\includegraphics[height=2.00cm]{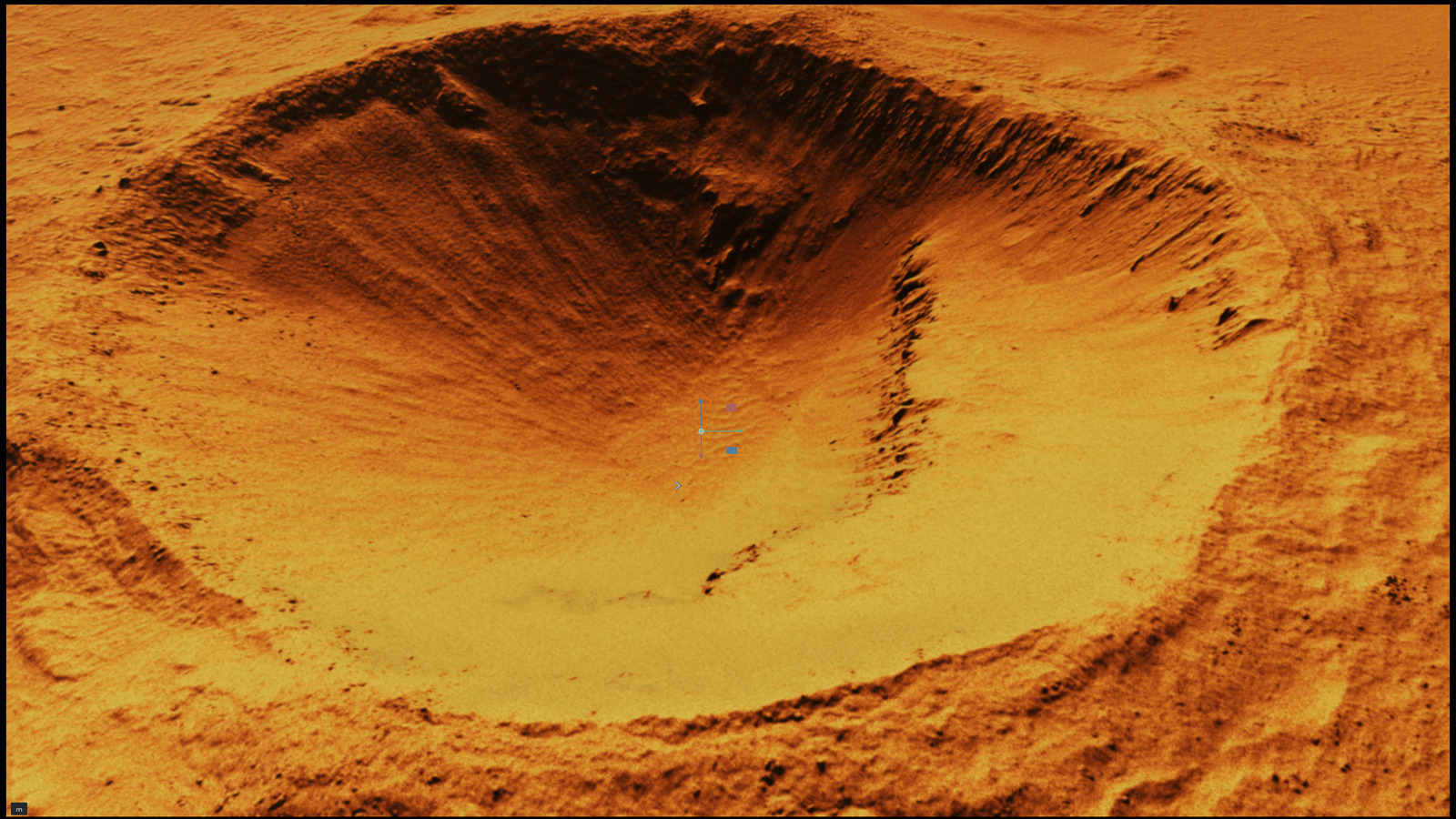}};
    \end{scope}
    \begin{scope}
      \clip[rounded corners=1.4pt] (\xthree+1.86,3.05) rectangle ++(1.09,0.61);
      \node[anchor=center, inner sep=0pt] at (\xthree+2.405,3.355)
        {\includegraphics[width=1.09cm,height=0.61cm,keepaspectratio]{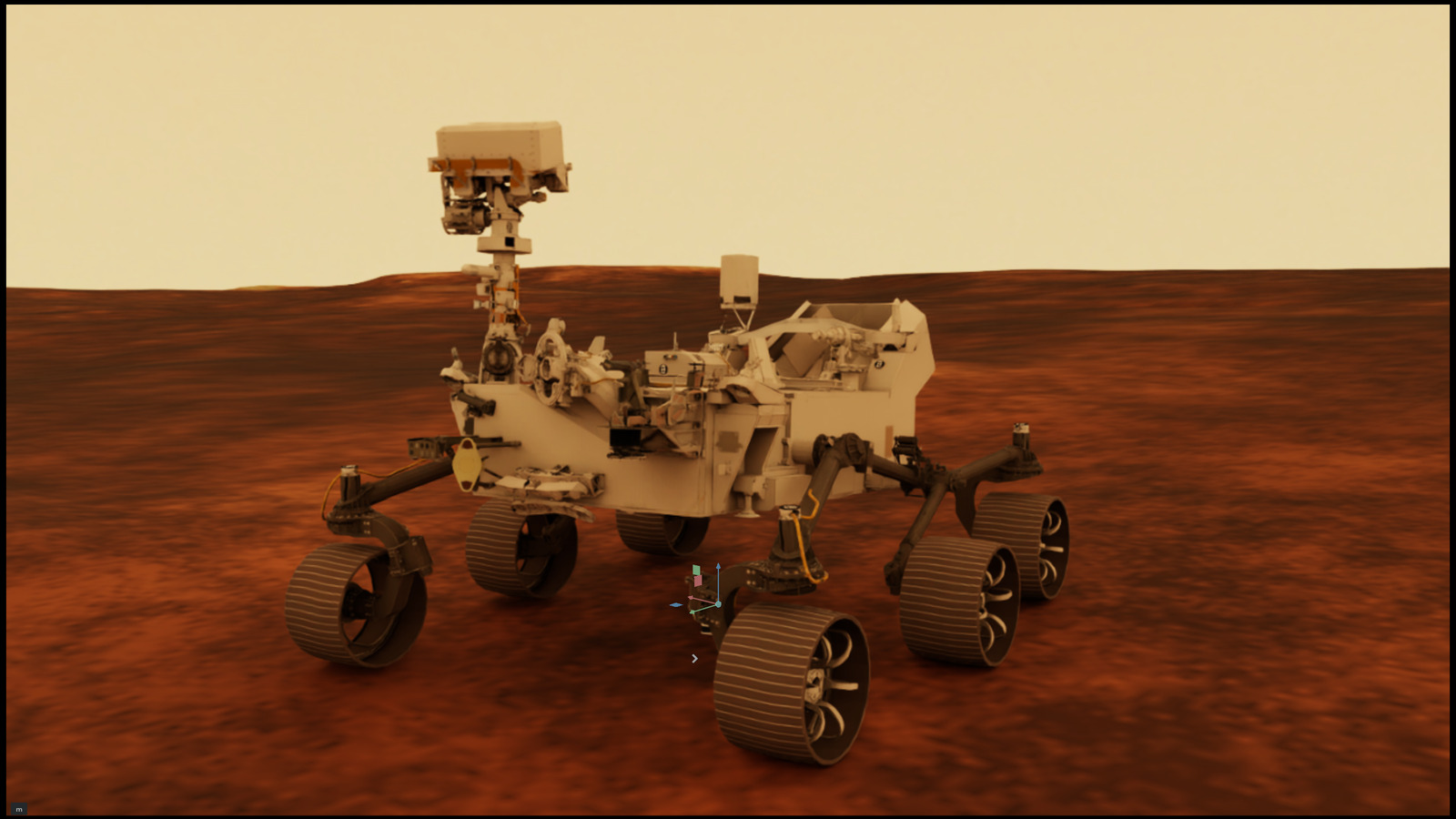}};
    \end{scope}
    \draw[rounded corners=1.4pt, draw=white, line width=0.75pt] (\xthree+1.86,3.05) rectangle ++(1.09,0.61);
    \node[label, anchor=center, align=center] at (\xthree+1.775,3.98) {MarsLab Runtime Scene};

    \foreach \name/\file/\xx in {RGB/fig02_rgb/0.24, Depth/fig02_depth_continuous/1.30}{%
      \node[text=mlpmuted, font=\sffamily\fontsize{4.7}{5.5}\selectfont\bfseries, anchor=south] at (\xthree+\xx+0.47,4.53) {\name};
      \begin{scope}
        \clip[rounded corners=1.8pt] (\xthree+\xx,4.53) rectangle ++(0.94,0.78);
        \node[anchor=center, inner sep=0pt] at (\xthree+\xx+0.47,4.92)
          {\includegraphics[width=0.94cm,height=0.78cm,keepaspectratio]{fig02/\file}};
      \end{scope}
      \draw[imageframe] (\xthree+\xx,4.53) rectangle ++(0.94,0.78);
    }
    \node[text=mlpmuted, font=\sffamily\fontsize{4.7}{5.5}\selectfont\bfseries, anchor=south] at (\xthree+2.83,4.53) {LiDAR};
    \begin{scope}
      \clip[rounded corners=1.8pt] (\xthree+2.36,4.53) rectangle ++(0.94,0.78);
      \node[anchor=center, inner sep=0pt] at (\xthree+2.83,4.92)
        {\includegraphics[height=0.78cm]{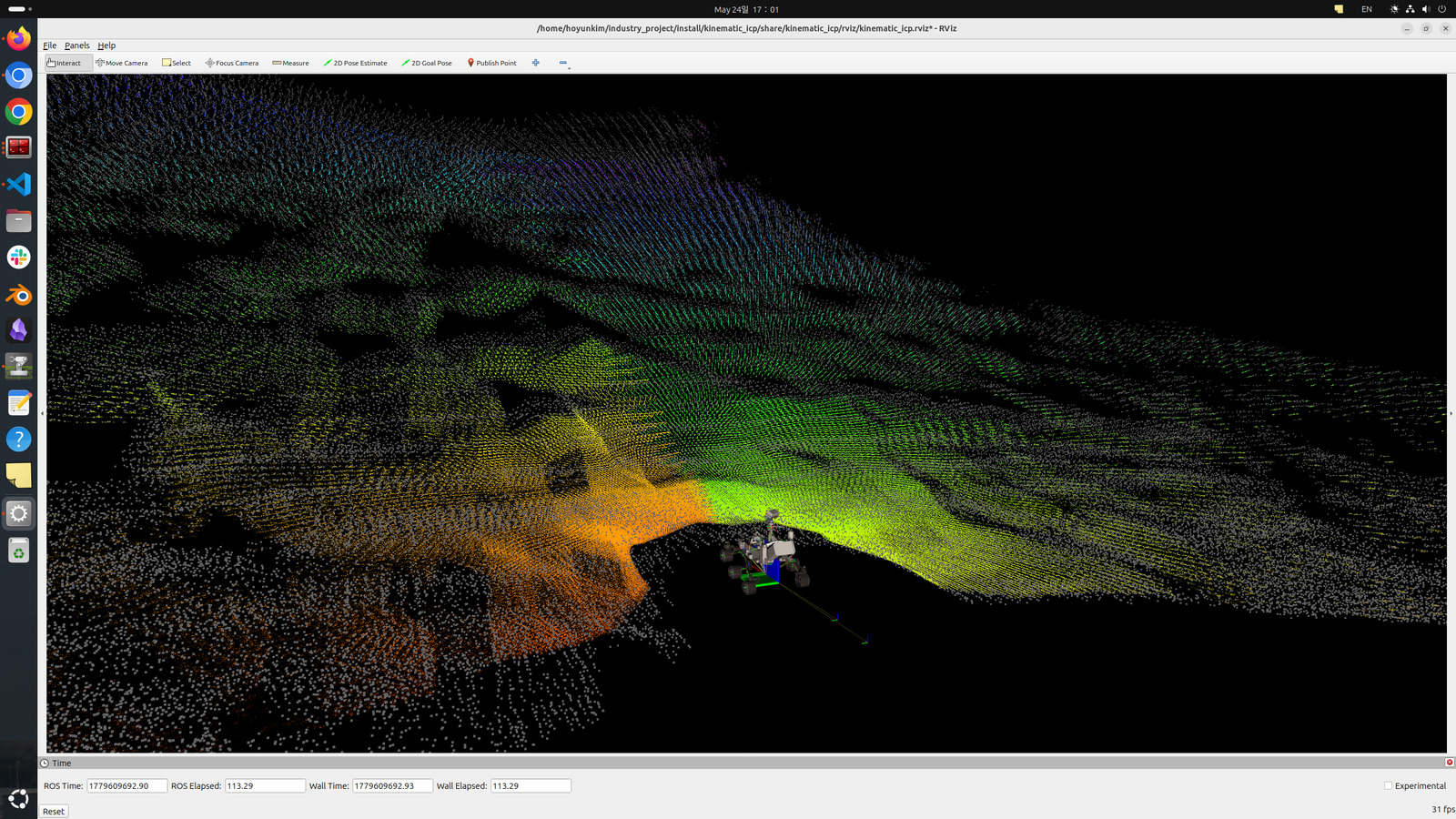}};
    \end{scope}
    \draw[imageframe] (\xthree+2.36,4.53) rectangle ++(0.94,0.78);
    \draw[rounded corners=2.1pt, draw=mlpline, line width=0.35pt, fill=white] (\xthree+0.38,5.37) rectangle ++(1.20,0.40);
    \node[text=mlptext, font=\sffamily\fontsize{4.8}{5.5}\selectfont\bfseries, anchor=center, align=center]
      at (\xthree+0.98,5.57) {IMU};
    \draw[rounded corners=2.1pt, draw=mlpline, line width=0.35pt, fill=white] (\xthree+1.75,5.37) rectangle ++(1.42,0.40);
    \node[text=mlptext, font=\sffamily\fontsize{4.5}{4.9}\selectfont\bfseries, anchor=center, align=center, text width=1.32cm]
      at (\xthree+2.46,5.57) {Wheel\\Odometry};
    \node[text=mlptext, font=\sffamily\fontsize{5.8}{6.7}\selectfont\bfseries, anchor=north, align=center]
      at (\xthree+1.775,5.79) {Sensing Modality};
    \draw[rounded corners=3pt, draw=mlporange, line width=0.45pt, fill=white] (\xfour+0.34,1.10) rectangle ++(2.86,0.56);
    \node[text=mlptext, font=\sffamily\fontsize{6.2}{7.2}\selectfont\bfseries, anchor=center, align=center] at (\xfour+1.77,1.38) {GT Pose};
    \foreach \txt/\yy/\col in {
      SLAM/2.18/mlpblue,
      Localization/2.98/mlpteal,
      Place Recognition/3.78/mlporange,
      Path Planning/4.58/mlpblue,
      Navigation/5.38/mlpteal}{%
      \draw[rounded corners=3pt, draw=\col, line width=0.45pt, fill=white] (\xfour+0.34,\yy) rectangle ++(2.86,0.46);
      \node[text=mlptext, font=\sffamily\fontsize{6.2}{7.2}\selectfont\bfseries, anchor=center, align=center] at (\xfour+1.77,\yy+0.23) {\txt};
    }
    \draw[arrow] (\xfour+1.77,1.66) -- (\xfour+1.77,2.18);
    \draw[arrow] (\xone+\cw,3.30) -- (\xtwo,3.30);
    \draw[arrow] (\xtwo+\cw,3.30) -- (\xthree,3.30);
    \draw[arrow] (\xthree+\cw,3.30) -- (\xfour,3.30);

  \end{tikzpicture}}
  \caption{\textbf{MarsLab pipeline.} MarsLab converts planetary terrain,
  surface assets, and rover models into customizable Mars experiments.  The
  pipeline builds scene variants with controlled terrain generation, rocks,
  craters, illumination, and dust; executes the rover in an Isaac~Sim/ROS2
  runtime under Mars gravity; and exports sensor, command, and GT pose
  for SLAM, localization, place recognition, planning, and navigation
  evaluation.}
  \label{fig:pipeline}
\tightfloatvspace
\end{figure*}

\section{Related Work}
\label{sec:related}

\subsection{Planetary Simulation and Autonomy Validation}
Simulation is an established approach for planetary-surface validation, but
existing Mars tools remain limited in scope and availability.  ROAMS and the
Mars 2020 ENav simulator are
mission-oriented and not community benchmarks~\cite{roams,toupet2020enavsim}.
ROS/Gazebo and Isaac~Sim systems such as Giubilato et
al.~\cite{giubilato2020planetarysim}, MarsSim~\cite{jiang2022marssim},
ISMRS~\cite{wan2024ismrs}, and RLRoverLab~\cite{mortensen2024rlroverlab}
address \ac{SLAM} evaluation, visual/physical fidelity, digital twins, or \ac{RL}, but do
not jointly provide an open ROS2 benchmark with customizable Mars terrain,
illumination, dust, sensors, and ground truth poses.  Table~\ref{tab:mars-sim-survey}
positions MarsLab against this landscape.

\subsection{Planetary Datasets, Semantics, and Traversability}
Planetary datasets capture important perceptual information, but their recorded
trajectories, environmental conditions, and sensor configurations are typically
fixed after release.
AI4MARS supplies terrain labels for Mars imagery~\cite{swan2021ai4mars}.
The Canadian energy-aware navigation dataset provides rover-collected
multi-sensor logs~\cite{lamarre2020canadian}, and DLR S3LI provides handheld
stereo, solid-state LiDAR, and inertial recordings from a planetary-analog
environment~\cite{giubilato2022s3li}.
They support offline perception studies, but cannot vary GT pose trajectories, terrain
statistics, dust, illumination, or sensor suites to measure closed-loop effects.

\subsection{Long-Term Localization and Mars \texorpdfstring{\ac{SLAM}}{SLAM}}
Long-term localization is central to planetary autonomy because drift propagates
to mapping, traversability assessment, and planning.  The Mars Exploration
Rovers have used visual odometry~\cite{maimone2007twoyearsvo}.
Orbital-map matching for global correction has been evaluated in terrestrial
analog terrain~\cite{geromichalos2020slam}, Censible has demonstrated
onboard global localization on Perseverance in shadow mode~\cite{nash2024censible}.
Direct validation on Mars remains limited by cost, sparse ground truth pose
data, and non-repeatable conditions.

MarsLab addresses this gap with HiRISE-derived and procedural terrain, a
Perseverance-class rover, controlled illumination and dust, synchronized ground
truth poses, and a standardized ROS2 benchmark.

\section{MarsLab Architecture}
\label{sec:system}

\subsection{System Overview}
\label{sec:overview}

MarsLab is an open-source, ROS2-native simulator built on NVIDIA Isaac
Sim~5.1~\cite{nvidia2024isaacsim}.  As shown in \figref{fig:pipeline}, the
pipeline starts from reusable terrain, surface, and rover assets; constructs
customizable Mars scenes; runs them with Isaac~Sim physics and ROS2 streams; and
exports experiment data for evaluation.  Each run is specified by a \ac{USD} scene,
YAML runtime configuration, seed, and trajectory/ground-truth pose data.  Runtime
parameters are validated before launch, and randomized stages such as crater
sampling, rock placement, and sensor noise share the explicit seed for
reproducible replay.

The architecture is organized as an end-to-end autonomy-evaluation pipeline.
The scene-construction layer builds Mars-relevant worlds from terrain sources
and customizable surface and environmental parameters.  The rover-runtime layer
loads each world into Isaac~Sim, applies Mars gravity, and provides rover
dynamics, command topics, sensor streams, and coordinate frames.  The benchmark
layer exports \ac{GT} trajectory and sensor data for evaluating \ac{SLAM}, localization,
navigation, \ac{VPR}, and related autonomy algorithms.  This separation lets users
vary scene conditions without changing
the rover interface or the evaluation format.

\begin{figure*}[!t]
  \centering
  \setlength{\tabcolsep}{1.2pt}%
  \renewcommand{\arraystretch}{1.0}%
  \newcommand{\scw}{0.242\textwidth}%
  \newcommand{\sceneviewbox}[5]{%
    \begin{tikzpicture}
      \node[anchor=south west, inner sep=0pt] (img) at (0,0)
        {\includegraphics[width=\scw]{#1}};
      \begin{scope}[x={(img.south east)}, y={(img.north west)}]
        \draw[green!75!black, line width=0.9pt,
          preaction={draw=black, line width=1.5pt, opacity=0.45}]
          (#2,#3) rectangle (#4,#5);
      \end{scope}
    \end{tikzpicture}%
  }%
  \begin{tabular}{@{}cccc@{}}
    \sceneviewbox{fig03/fig03_jezero1}{0.43}{0.33}{0.62}{0.58} &
    \sceneviewbox{fig03/fig03_main_crater1}{0.31}{0.21}{0.58}{0.49} &
    \sceneviewbox{fig03/fig03_marsbase1}{0.50}{0.34}{0.70}{0.58} &
    \sceneviewbox{fig03/fig03_marscanyon1}{0.54}{0.20}{0.80}{0.55} \\
    {\footnotesize Jezero plain} & {\footnotesize Main Crater} &
    {\footnotesize Mars Base} & {\footnotesize Grand Canyon} \\[3pt]
    \includegraphics[width=\scw]{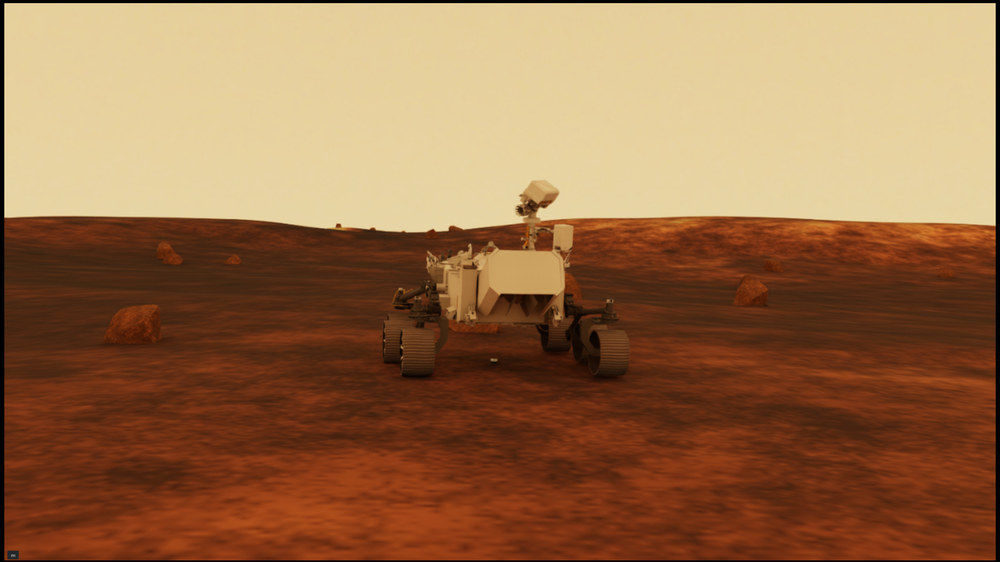} &
    \includegraphics[width=\scw]{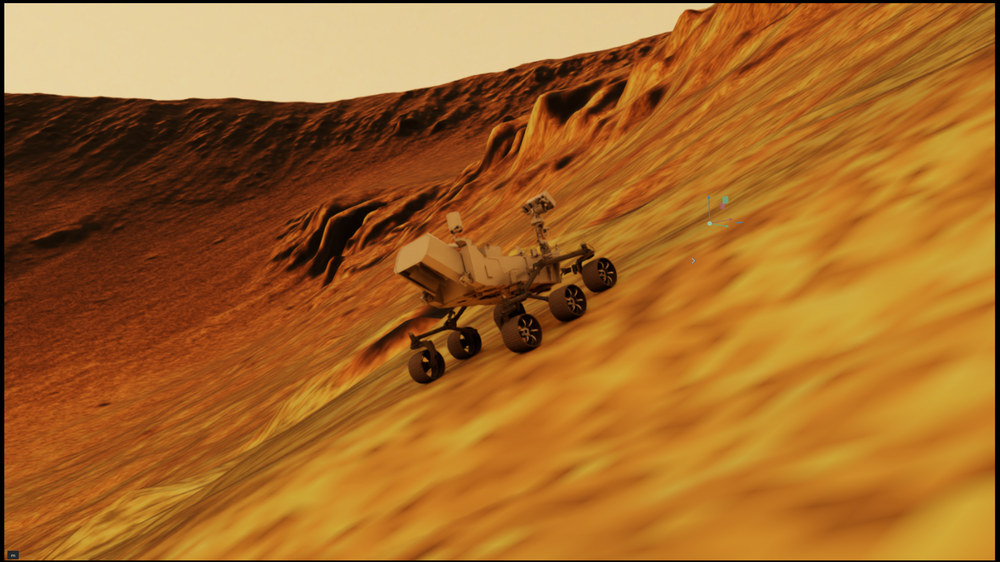} &
    \includegraphics[width=\scw]{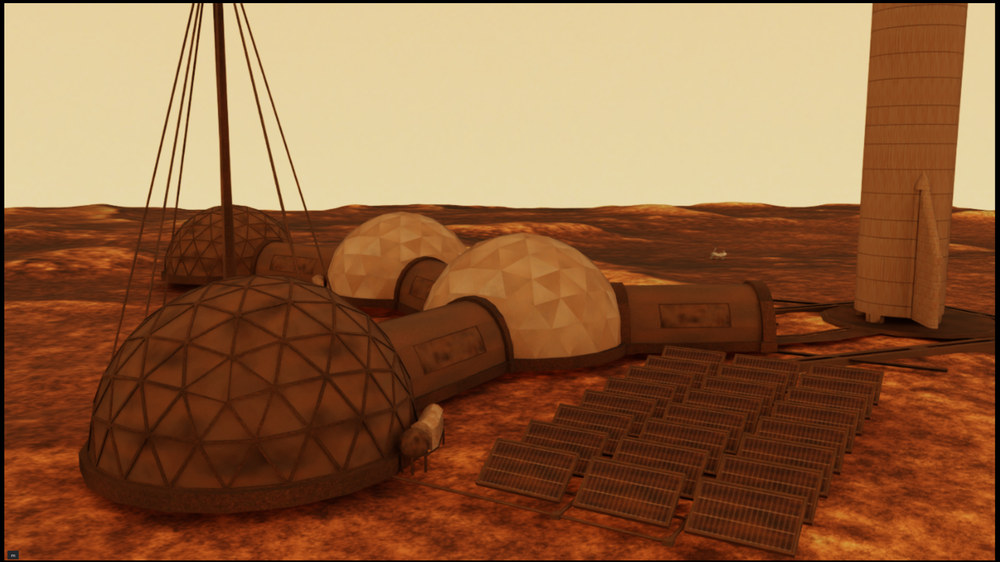} &
    \includegraphics[width=\scw]{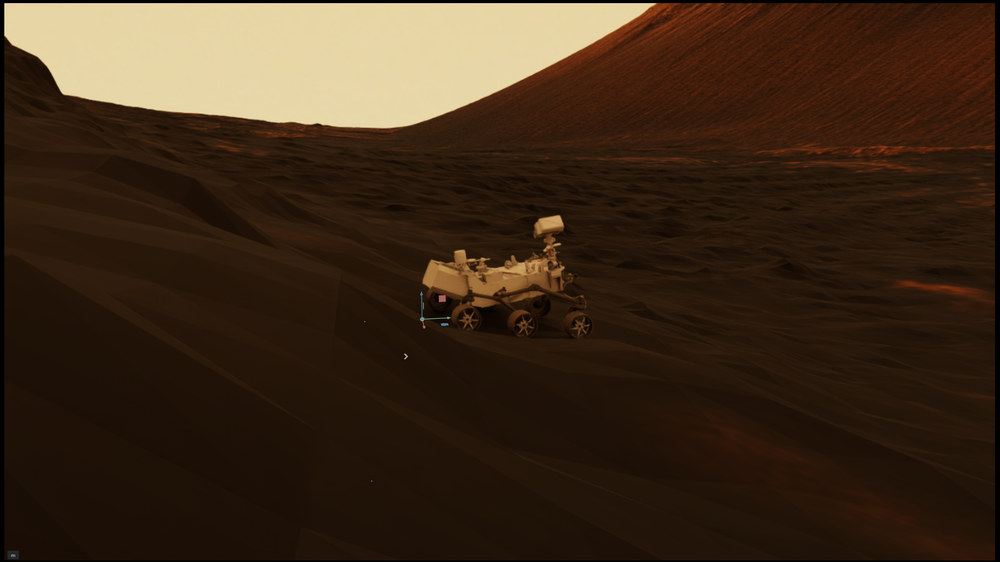} \\
    {\footnotesize Jezero rim} & {\footnotesize Crater interior} &
    {\footnotesize Base assets} & {\footnotesize Canyon wall} \\[-1pt]
    {\footnotesize (a)} & {\footnotesize (b)} & {\footnotesize (c)} & {\footnotesize (d)} \\
  \end{tabular}
  \caption{\textbf{Reference MarsLab scenes.} MarsLab can generate additional
  terrain instances, and these four reference scene pairs are provided as
  scenes for evaluation.  The green rectangle in each upper image indicates the
  approximate area represented by the corresponding lower scene: (a) a plain
  scene from Jezero crater, (b) a main-crater scene from Arcadia Planitia,
  (c) a Mars Base scene that augments the Jezero terrain with a habitat asset,
  and (d) a Grand Canyon scene from Elysium Planitia.  All scenes are built on
  HiRISE-derived terrain, providing diverse geometry, texture, and structure for
  SLAM and place recognition stress testing.}
  \label{fig:scenes}
\tightfloatvspace
\end{figure*}
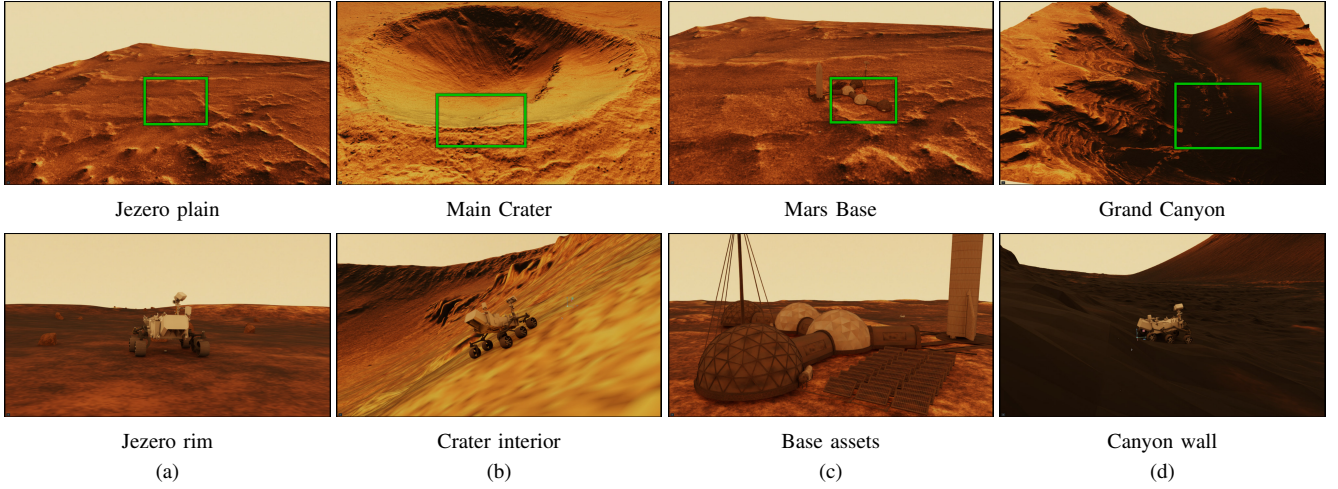

\begin{table*}[!t]
  \centering
  \caption{\textbf{MarsLab reference scenes.} Compact summary of the four evaluation scenes in \figref{fig:scenes}.  Rock distribution, illumination, and atmospheric dust denote user-customizable scene parameters, including rock size--frequency/spatial-density settings, solar azimuth/elevation, and dust optical depth.}
  \label{tab:reference-scenes}
  \footnotesize
  \setlength{\tabcolsep}{1.8pt}%
  \renewcommand{\arraystretch}{1.58}%
  \renewcommand{\tabularxcolumn}[1]{m{#1}}%
  \begin{tabularx}{\textwidth}{@{}>{\centering\arraybackslash}m{0.035\textwidth}>{\raggedright\arraybackslash}m{0.145\textwidth}>{\raggedright\arraybackslash}m{0.095\textwidth}>{\centering\arraybackslash}m{0.045\textwidth}>{\centering\arraybackslash}m{0.045\textwidth}>{\centering\arraybackslash}m{0.085\textwidth}@{\hspace{1.1em}}>{\raggedright\arraybackslash}X@{}}
    \toprule
    Fig. & Reference scene & Scene size & Rock & Dust & Illumination & Target Use \\
    \midrule
    (a) & \textbf{Jezero plain/rim} & $0.5{\times}0.5$\,km & \yes & \yes & \yes & Navigation, terrain following, and low-texture localization baselines. \\
    (b) & \textbf{Main Crater} & $3.6{\times}3.6$\,km & \yes & \yes & \yes & Repetitive-terrain failure analysis and rocky slope traversal. \\
    (c) & \textbf{Mars Base} & $1.0{\times}1.0$\,km & \yes & \yes & \yes & Landmark- and feature-rich localization and VPR benchmarking. \\
    (d) & \textbf{Grand Canyon} & $3.0{\times}3.0$\,km & \yes & \yes & \yes & Long-range drift evaluation over extended canyon terrain with strong elevation changes and complex geometry. \\
    \bottomrule
  \end{tabularx}
\tightfloatvspace
\end{table*}

\subsection{Terrain and Scene Generation}
\label{sec:scene}

MarsLab combines real and procedural terrain sources.  HiRISE \ac{DEM}s and
co-registered orthomosaics provide real terrain geometry and texture information, with
Mastcam-Z imagery as a surface-appearance reference~\cite{mcewen2007hirise,Hayes2021}.
Procedural terrains are synthesized with \texttt{synthterrain}~\cite{neogeography2023synthterrain}
and augmented with the rock and crater models below.  The reference scenes span
plains, craters, canyons, and habitat sites%
\footnote{``Mars Base'' by MOJackal,
  via Sketchfab,
  \url{https://sketchfab.com/3d-models/mars-base-2bc6660c63b24158bdded20ff5d2235e},
  licensed under CC~BY~4.0.\label{fn:mars-base}} (\figref{fig:scenes}; Table~\ref{tab:reference-scenes}), while
\figrangeref{fig:scenes-proc}{fig:dust} show the customizable terrain,
illumination, and dust factors used as controlled evaluation conditions.

Scene construction follows a deterministic asset-build process.  Given a crop
region, terrain scale, and generation seed, the builder places rocks, crater
fields, rover assets, lighting, and atmospheric settings into a single world
description.  The generated scene is stored separately from the runtime
configuration, so it can be replayed with different sensors or controllers and
the same \ac{GT} trajectory can be evaluated under modified lighting or dust.  This
design keeps MarsLab from being a fixed dataset: it is a controlled generator of
related experiments whose differences are known before evaluation.

\begin{table}[t]
  \centering
  \caption{\textbf{Customizable MarsLab scene generation.} Beyond the reference
  scenes, MarsLab lets users build HiRISE-based scenes by controlling terrain
  extent, surface clutter, crater fields, illumination, and atmospheric dust.
  The table summarizes the scene-generation factors exposed as user-controlled
  variables.}
  \label{tab:scene-generation-factors}
  \footnotesize
  \setlength{\tabcolsep}{2.2pt}%
  \renewcommand{\arraystretch}{1.18}%
  \begin{tabularx}{\columnwidth}{@{}>{\raggedright\arraybackslash}p{0.25\columnwidth}>{\raggedright\arraybackslash}p{0.31\columnwidth}>{\raggedright\arraybackslash}X@{}}
    \toprule
    \textbf{Stage} & \textbf{User-provided input} & \textbf{Customizable variation} \\
    \midrule
    Terrain base &
    HiRISE DEM/orthomosaic &
    Crop region, terrain scale, and generation seed. \\
    Surface clutter &
    Rock assets and rock size--frequency parameters &
    Rock density, diameter range, spatial sampling, and placement seed. \\
    Crater field &
    Mars crater production-function parameters &
    Crater density, diameter range, morphology, and resurfacing masks. \\
    Illumination &
    Solar-position and irradiance configuration &
    Sun azimuth, elevation, shadow direction, and scene shading. \\
    Atmosphere &
    Mars dust optical-depth setting &
    Dust optical depth $\tau$, sky brightness, and image contrast. \\
    \bottomrule
  \end{tabularx}
\tightfloatvspace
\end{table}

\subsection{Rock Distribution}
\label{sec:rock}

Rock populations follow the Golombek--Huertas cumulative fractional-area
model~\cite{golombek2008northernplains}:
\begin{equation}
  F_k(D) = k\,e^{-q(k)\,D},
  \qquad q(k) = 1.79 + \frac{0.152}{k},
  \label{eq:rock-sfd}
\end{equation}
where $D$ is rock diameter, $k$ is the total rock-covered area fraction, and
$F_k(D)$ is the cumulative area fraction covered by rocks with diameter at least
$D$.  Sampled diameters are used to scale \ac{USD} rock assets%
\footnote{``Mars Rocks'' by Ivan Vakulko
  (milos4), via Sketchfab,
  \url{https://sketchfab.com/3d-models/mars-rocks-9f5c946255a24f1cb630ea95dceea587},
  licensed under CC~BY~4.0.\label{fn:mars-rocks}} on the \ac{DEM} surface, so the
generated rock field follows the target size--frequency statistics rather than
arbitrary placement (\figref{fig:rock}).

\subsection{Crater Distribution}
\label{sec:crater}

\begin{figure}[tbp]
  \centering
  \vspace{-0.35em}
  \begin{subfigure}[t]{0.49\linewidth}
    \centering
    \includegraphics[width=\linewidth,height=0.108\textheight,keepaspectratio]{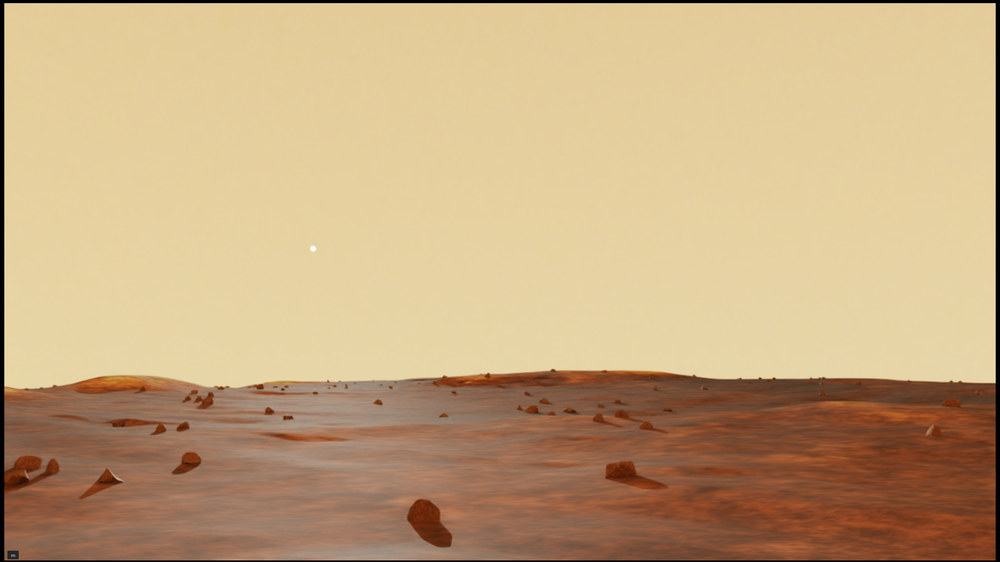}
    \caption{Rocky scene.}
    \label{fig:rock}
  \end{subfigure}\hfill
  \begin{subfigure}[t]{0.49\linewidth}
    \centering
    \includegraphics[width=\linewidth,height=0.108\textheight,keepaspectratio]{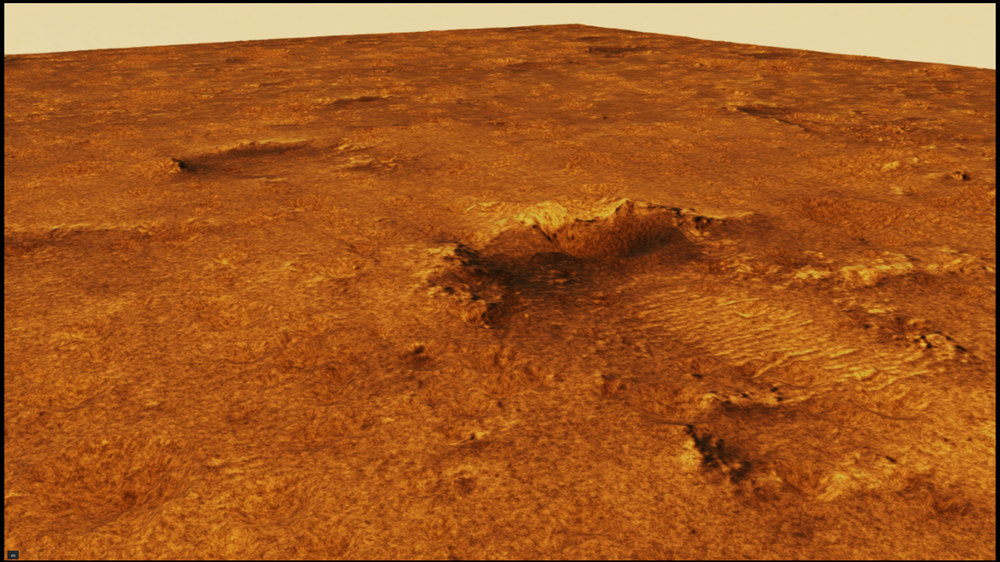}
    \caption{Crater scene.}
    \label{fig:crater}
  \end{subfigure}
  \caption{\textbf{Procedurally generated rock and crater scenes.} Rendered MarsLab terrain showing (a) the rock size--frequency distribution (\eqnref{eq:rock-sfd}) and (b) crater density and morphometry (\eqnref{eq:crater-cdf}).}
  \label{fig:scenes-proc}
\tightfloatvspace
\end{figure}

Crater diameters are sampled from the Hartmann--Neukum Mars production
function~\cite{hartmann2001chronology}.  For the valid $D\geq10$\,m regime,
MarsLab converts the cumulative SFD $C(D)$ into a normalized sampling CDF:
\begin{equation}
  F(D) = 1 - \frac{C(D)}{C(D_{\min})},
  \qquad D\in[D_{\min},D_{\max}].
  \label{eq:crater-cdf}
\end{equation}
Optional resurfacing masks follow crater-population clustering
analysis~\cite{michael2010dating}, and diameter-dependent profiles use published
fresh-crater morphometry~\cite{watters2015morphometry}; thus crater density and
shape remain tied to planetary chronology models (\figref{fig:crater}).

\subsection{Solar Illumination and Dust Environment}
\label{sec:atmosphere}

\begin{figure}[tbp]
  \centering
  \begin{subfigure}[t]{0.49\linewidth}
    \centering
    \includegraphics[width=\linewidth,height=0.135\textheight,keepaspectratio]{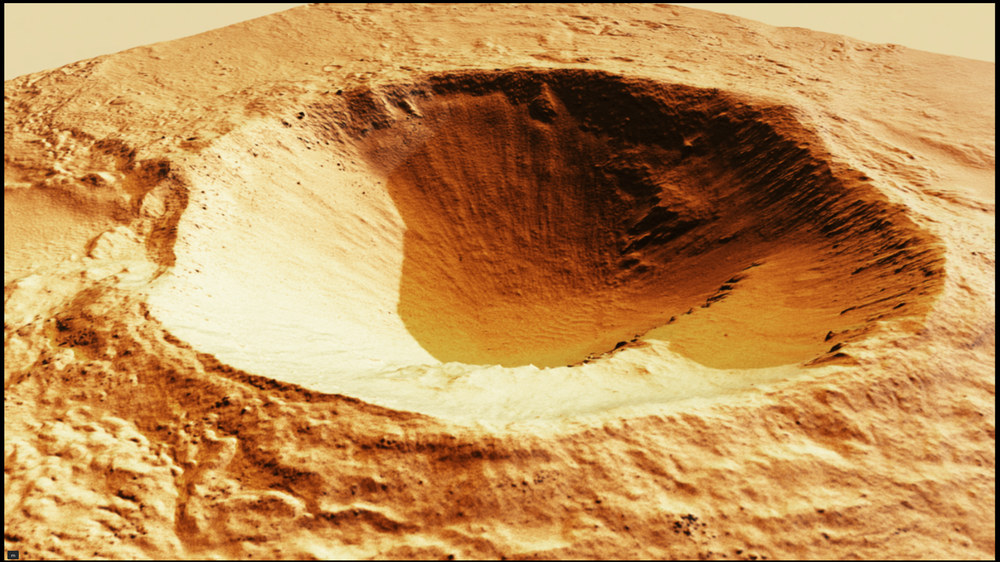}
    \caption{Azimuth $70^\circ$, elevation $13^\circ$}
  \end{subfigure}\hfill
  \begin{subfigure}[t]{0.49\linewidth}
    \centering
    \includegraphics[width=\linewidth,height=0.135\textheight,keepaspectratio]{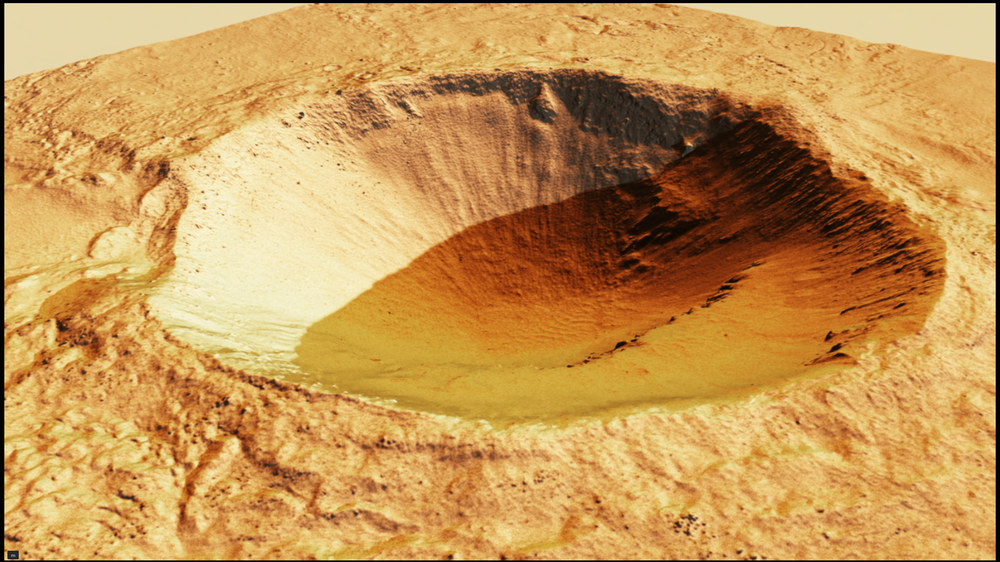}
    \caption{Azimuth $360^\circ$, elevation $13^\circ$}
  \end{subfigure}
  \caption{\textbf{Controllable solar illumination.} The sun azimuth and elevation are explicit experimental variables; changing the solar position redistributes cast shadows and shading across the same terrain, directly affecting feature visibility for vision-based navigation algorithms.}
  \label{fig:sun}
\tightfloatvspace
\end{figure}

\begin{figure}[tbp]
  \centering
  \begin{subfigure}[t]{0.32\linewidth}
    \centering
    \includegraphics[width=\linewidth,height=0.095\textheight,keepaspectratio]{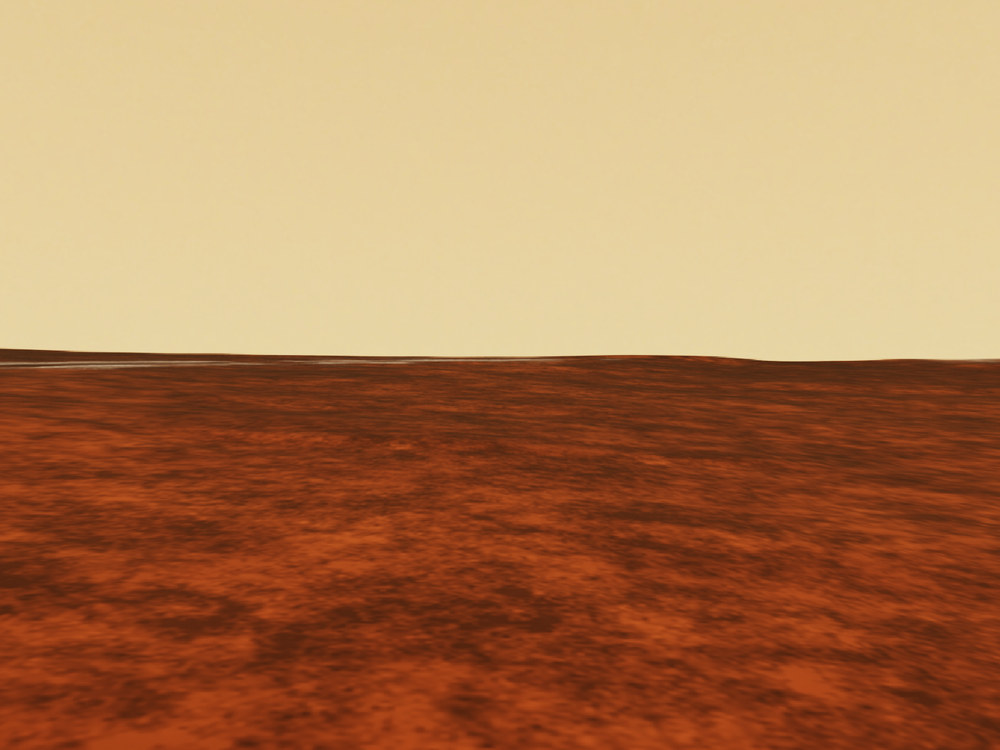}
    \caption{$\tau=0.3$}
  \end{subfigure}\hfill
  \begin{subfigure}[t]{0.32\linewidth}
    \centering
    \includegraphics[width=\linewidth,height=0.095\textheight,keepaspectratio]{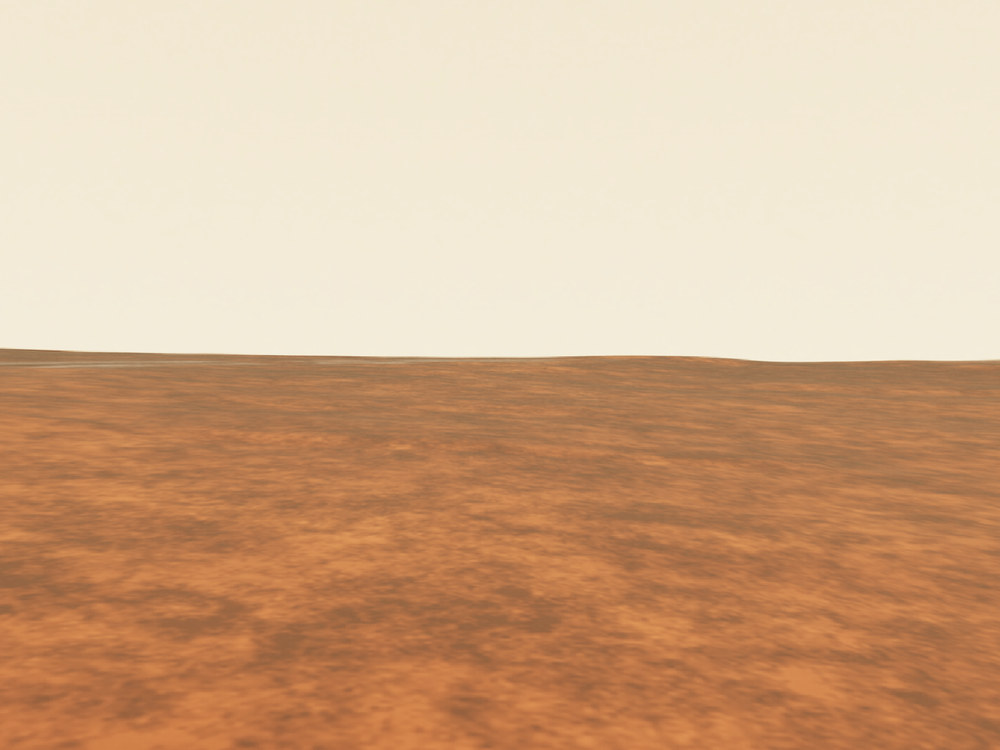}
    \caption{$\tau=3.0$}
  \end{subfigure}\hfill
  \begin{subfigure}[t]{0.32\linewidth}
    \centering
    \includegraphics[width=\linewidth,height=0.095\textheight,keepaspectratio]{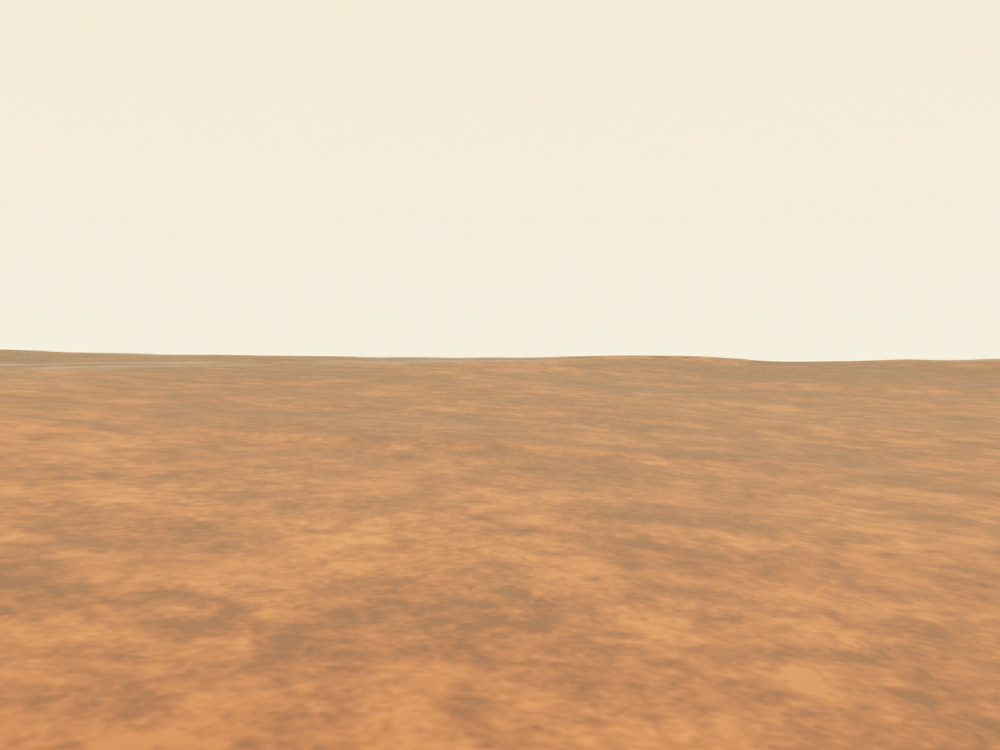}
    \caption{$\tau=6.0$}
  \end{subfigure}
  \caption{\textbf{Atmospheric-dust sweep.} MarsLab uses the optical-depth parameter $\tau$ in \eqnref{eq:beer} as a controlled evaluation condition for vision-based algorithms: increasing $\tau$ brightens the sky and ground, lowers image contrast, and reduces visibility while leaving the terrain geometry unchanged.}
  \label{fig:dust}
\tightfloatvspace
\end{figure}

Illumination is treated as a customizable experimental variable.  Direct
surface irradiance follows Beer's law with Mars dust optical depth $\tau$:
\begin{equation}
  I(z) =
  \begin{cases}
    I_0\,e^{-\tau\,m(z)}, & z < 90^\circ,\\[2pt]
    0, & z \geq 90^\circ,
  \end{cases}
  \label{eq:beer}
\end{equation}
where $z$ is solar zenith angle, $I_0\approx589\,\mathrm{W\,m^{-2}}$, and
$m(z)$ is the Appelbaum--Flood relative optical air mass~\cite{appelbaum1990solar}.
Diffuse sky illumination follows a $\tau$-indexed COMIMART
reduction~\cite{vicenteretortillo2015comimart}.  The same trajectory can then be
replayed with customizable solar azimuth/elevation and dust optical depth: sun
changes redistribute shadows (\figref{fig:sun}), while increasing $\tau$ reduces
contrast and visibility (\figref{fig:dust}) without changing geometry.
By varying one environmental parameter at a time, such as sun position or dust
optical depth, MarsLab enables vision-based algorithms to be tested under
diverse conditions within the same scene.

\definecolor{topicbg}{HTML}{F7F8FA}
\definecolor{topicline}{HTML}{C9CDD4}
\definecolor{topichead}{HTML}{2E3440}
\definecolor{topicpub}{HTML}{2F6F5E}
\definecolor{topicsub}{HTML}{8A4B2B}

\newcommand{\rostopicpub}[1]{\texttt{#1}{\color{topicpub}\texttt{[publish]}}}
\newcommand{\rostopicsub}[1]{\texttt{#1}{\color{topicsub}\texttt{[subscribe]}}}

\begin{figure}[tbp]
  \centering
  \setlength{\fboxsep}{0pt}%
  \fcolorbox{topicline}{topicbg}{%
  \begin{minipage}{0.99\columnwidth}
    \vspace{0.55em}
    \hspace{0.60em}{\ttfamily\scriptsize\bfseries\textcolor{topichead}{marslab@ros2:\string~\$ ros2 topic list}}
    \vspace{0.40em}

    \fontsize{5.5}{6.2}\selectfont
    \setlength{\tabcolsep}{1.2pt}%
    \renewcommand{\arraystretch}{1.24}%
    \hspace{0.60em}\begin{tabularx}{0.96\linewidth}{@{}>{\raggedright\arraybackslash}X>{\raggedright\arraybackslash}X@{}}
      \rostopicsub{/rover/cmd\_vel} & \rostopicpub{/rover/imu} \\
      \rostopicpub{/clock} & \rostopicpub{/rover/imu\_noisy} \\
      \rostopicpub{/tf} & \rostopicpub{/rover/rgb/image\_raw} \\
      \rostopicpub{/tf\_static} & \rostopicpub{/rover/rgb/camera\_info} \\
      \rostopicpub{/rover/joint\_states} & \rostopicpub{/rover/depth/image\_raw} \\
      \rostopicpub{/rover/robot\_description} & \rostopicpub{/rover/depth/points} \\
      \rostopicpub{/rover/odom} & \rostopicpub{/rover/lidar/points} \\
      \rostopicpub{/rover/GT\_Trajectory} & \rostopicpub{/rover/scan} \\
    \end{tabularx}
    \vspace{0.55em}
  \end{minipage}}
  \caption{\textbf{MarsLab ROS2 topic interface.} Default subscribed and published topics exposed by the rover runtime.}
  \label{fig:ros2-topics}
\tightfloatvspace
\end{figure}

\subsection{Rover Runtime and Physics}
\label{sec:rover}

The runtime uses a Perseverance (M2020) model from NASA-JPL's public
URDF release~\cite{nasajpl2022m2020urdf}, converted to \ac{USD} and simulated under
Mars gravity ($3.72\,\mathrm{m/s^2}$).  A sensor graph and \texttt{rclpy} bridge
subscribe to \texttt{cmd\_vel} and publish odometry, IMU, RGB, depth, RGB-D
point cloud, 2D/3D LiDAR, TF, robot description, and ground-truth poses in
REP-103/REP-105 frames (\figref{fig:ros2-topics}).  \ac{GT} trajectories are written
in TUM format and replayed by a pure-pursuit follower with bounded commands
($v\leq0.5\,\mathrm{m/s}$, $|\omega|\leq0.2\,\mathrm{rad/s}$, 50\,Hz), giving a
reference data stream for \ac{ATE}/RPE evaluation.  The resulting sensor streams and
\ac{GT} trajectory data provide the common interface used by the runtime and the
benchmarks below.
The runtime also supports experiments and analyses under varying sensor-noise
conditions by providing zero-mean Gaussian perturbations with customizable
standard deviations for IMU and wheel-odometry measurements, as well as optional
depth-camera post-processing with customizable noise mean and scale.

\newlength{\slamh}\setlength{\slamh}{0.175\textwidth}
\newcommand{\slamcell}[1]{\begin{minipage}[c][\slamh][c]{\slamh}\centering #1\end{minipage}}
\newcommand{\slamtile}[1]{\slamcell{\includegraphics[width=\slamh,height=\slamh,keepaspectratio]{fig08/#1_clean}}}
\newcommand{\slamfail}{\slamcell{{\setlength{\fboxsep}{0pt}\colorbox{gray!15}{%
  \parbox[c][\slamh][c]{0.72\slamh}{\centering\no\\[3pt]{\scriptsize tracking\\[-1pt] lost}}}}}}
\newcommand{\slamrlab}[1]{\begin{minipage}[c][\slamh][c]{14pt}\centering\rotatebox{90}{\small\textbf{#1}}\end{minipage}}

\newlength{\mapw}\setlength{\mapw}{0.305\textwidth}
\newlength{\maph}\setlength{\maph}{0.168\textwidth}
\newcommand{\mapcell}[1]{\begin{minipage}[c][\maph][c]{\mapw}\centering #1\end{minipage}}
\newcommand{\maptile}[1]{\mapcell{\includegraphics[width=\mapw,height=\maph,keepaspectratio]{fig8_new/#1_clean}}}
\newcommand{\maprlab}[1]{\begin{minipage}[c][\maph][c]{12pt}\centering\rotatebox{90}{\small\textbf{#1}}\end{minipage}}

\begin{figure*}[p]
  \centering
  \setlength{\tabcolsep}{0.8pt}%
  \renewcommand{\arraystretch}{0.95}%
  \begin{tabular}{@{}c@{\hspace{2pt}}*{5}{c}@{}}
    & \multicolumn{2}{c}{\textbf{ORB-SLAM (RGB)}}
    & \multicolumn{2}{c}{\textbf{RTAB-Map (RGB-D+Odom)}}
    & \textbf{MOLA (LiDAR)} \\
    \cmidrule(lr){2-3}\cmidrule(lr){4-5}\cmidrule(lr){6-6}
    & {\small$\tau{=}0.5$} & {\small$\tau{=}6.0$}
    & {\small$\tau{=}0.5$} & {\small$\tau{=}6.0$} & {\small nominal} \\
    \slamrlab{Mars Base}
      & \slamtile{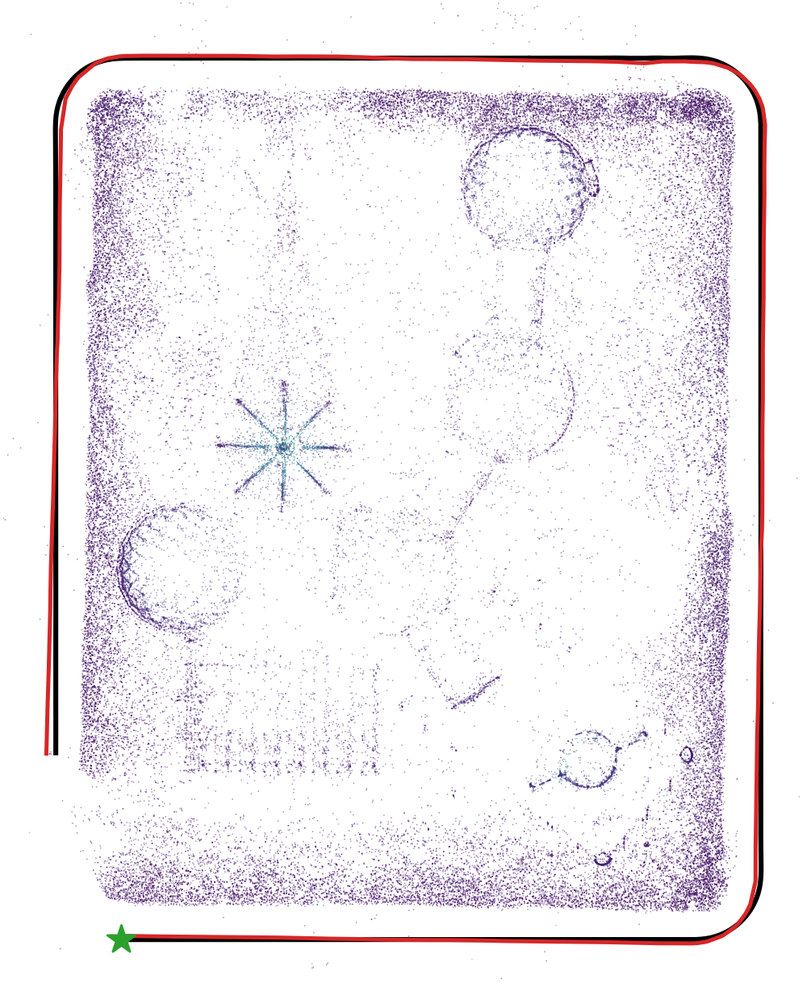} & \slamtile{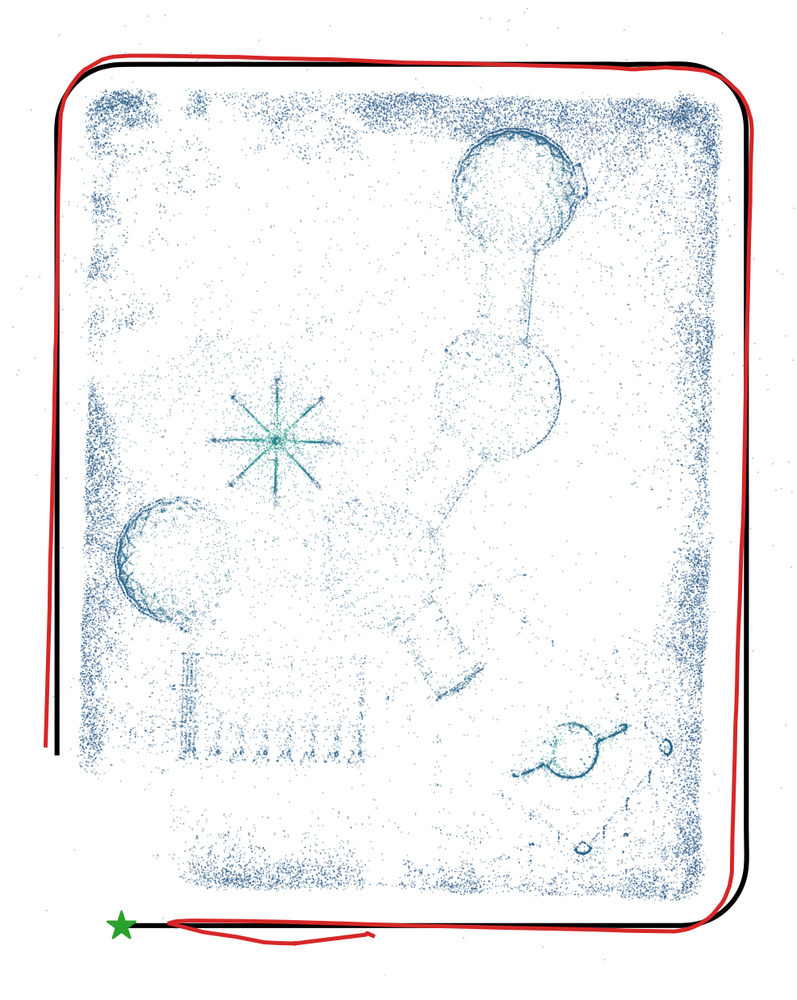}
      & \slamtile{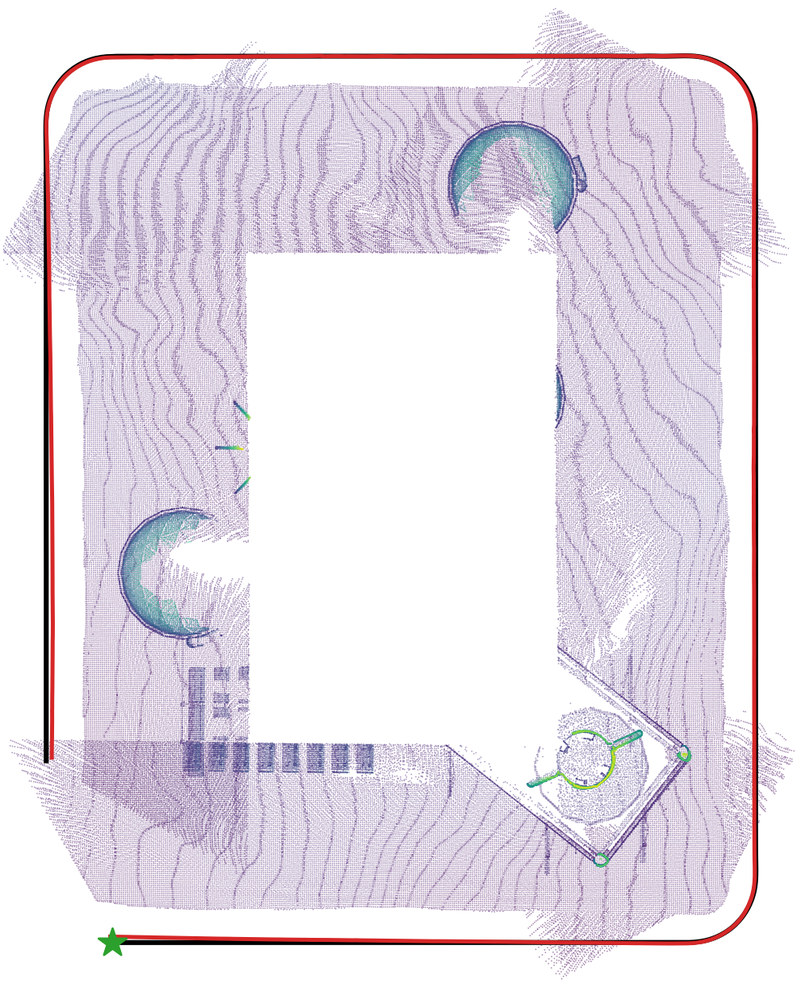} & \slamtile{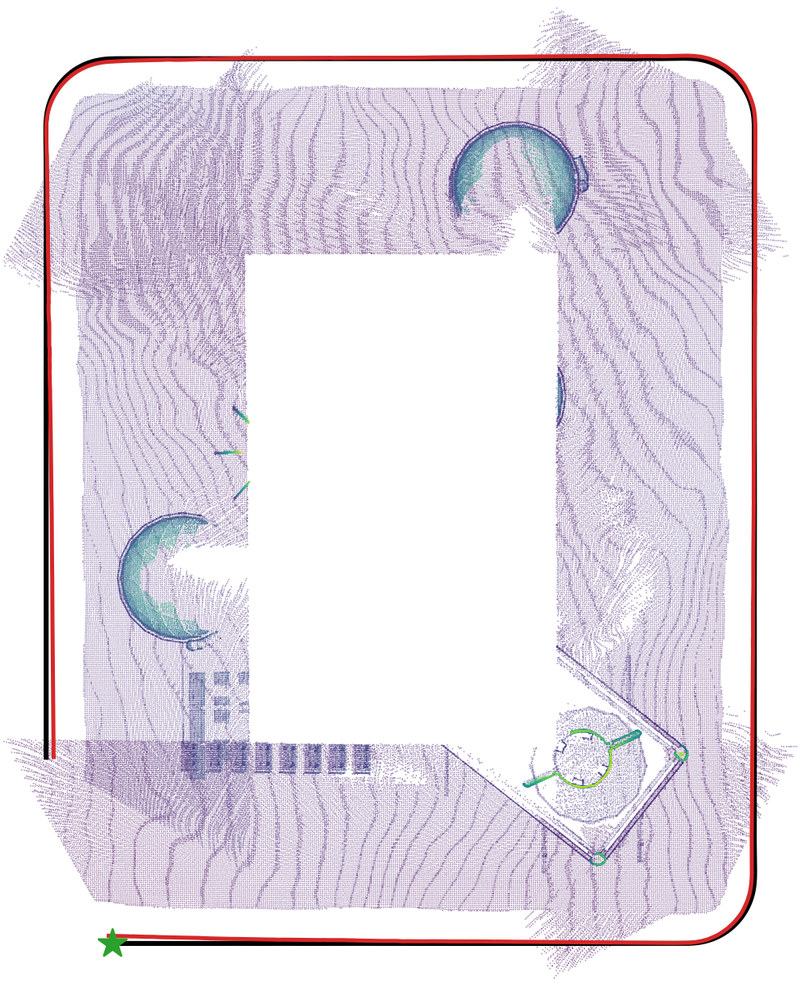}
      & \slamtile{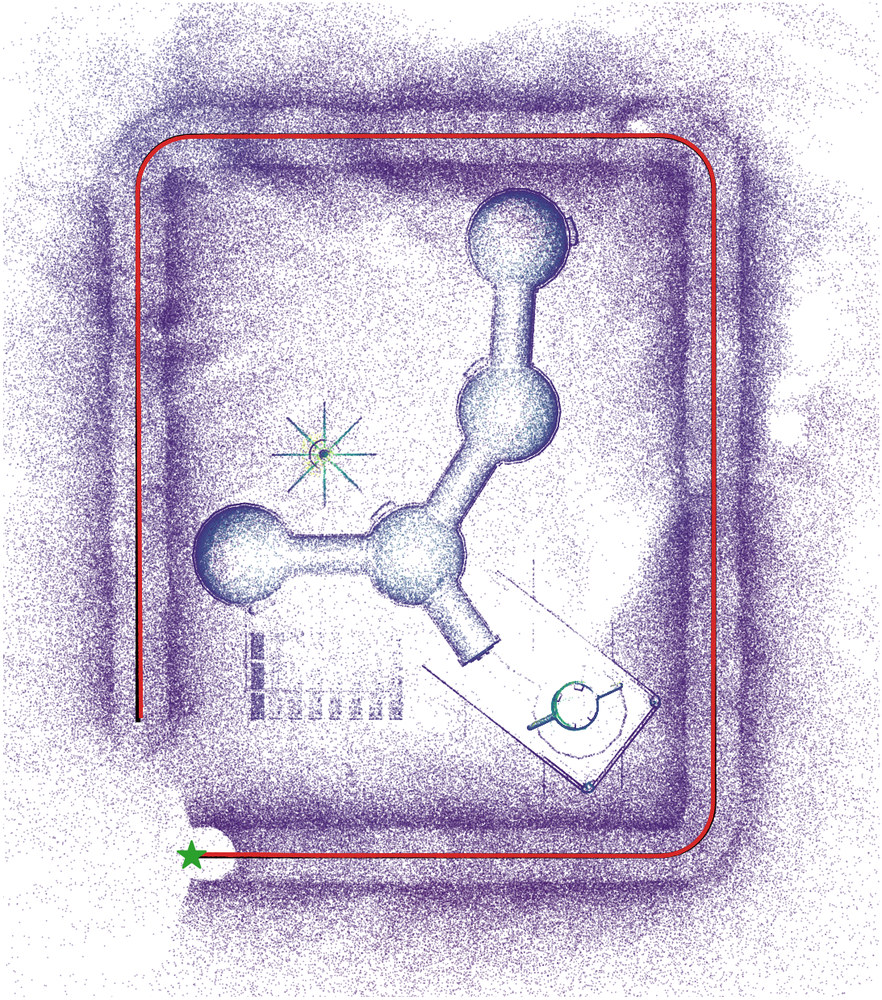} \\
    \slamrlab{Main Crater Interior}
      & \slamtile{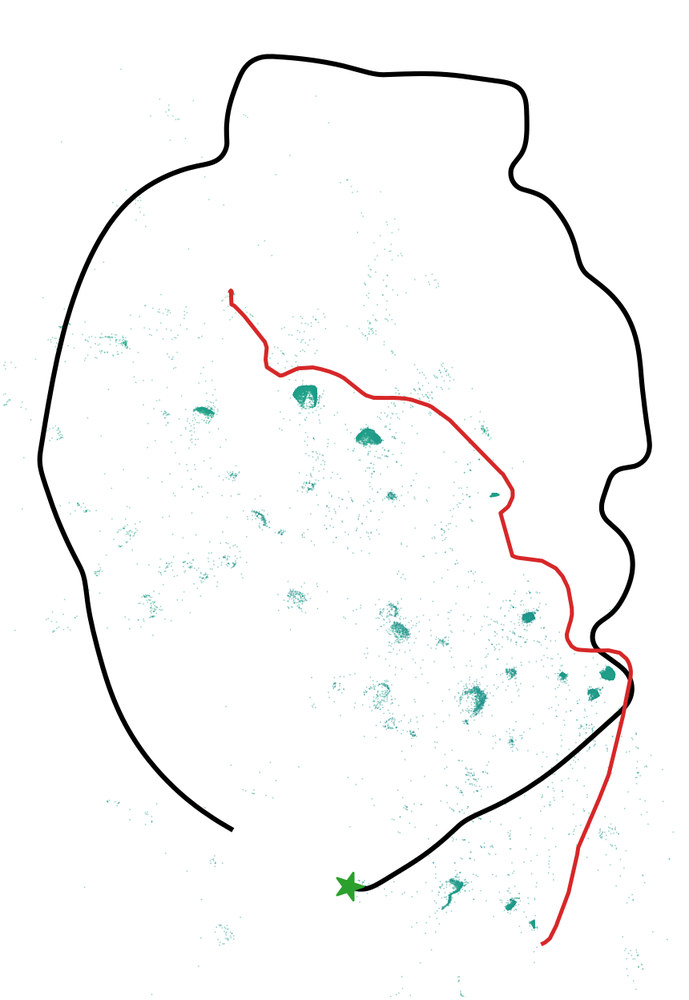} & \slamfail
      & \slamtile{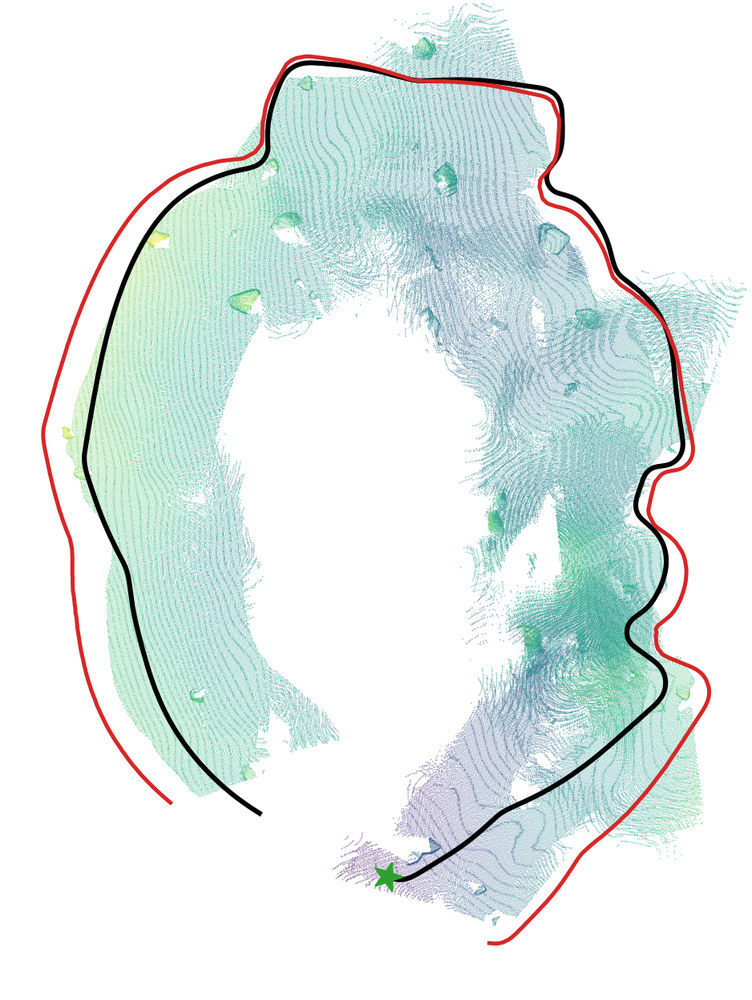} & \slamtile{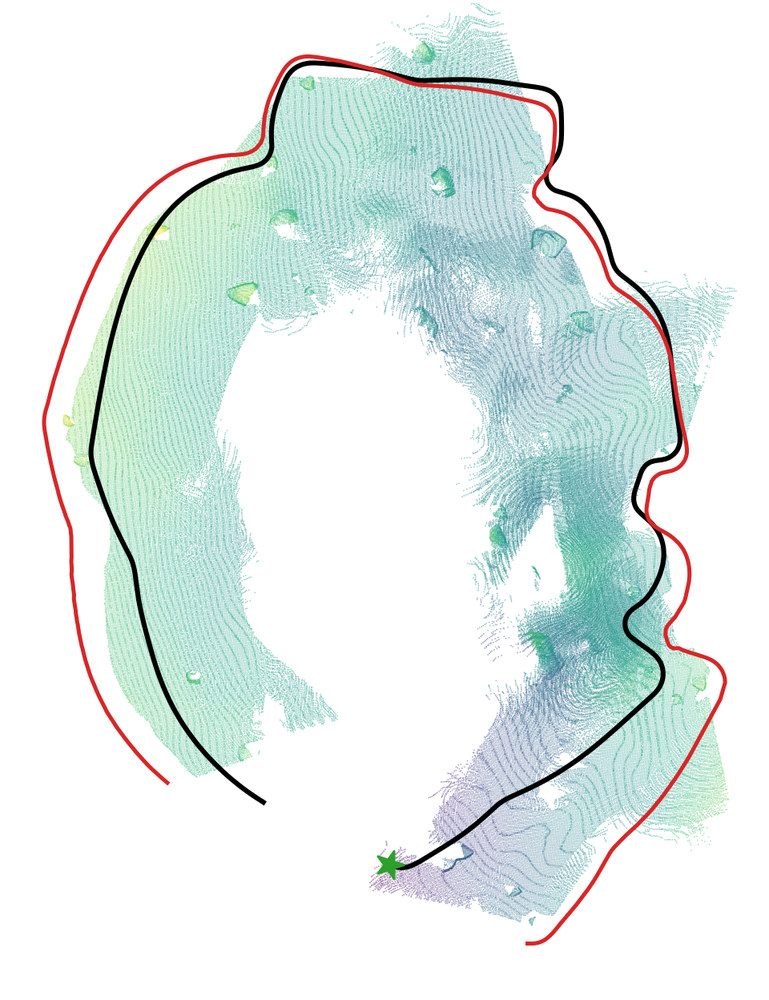}
      & \slamtile{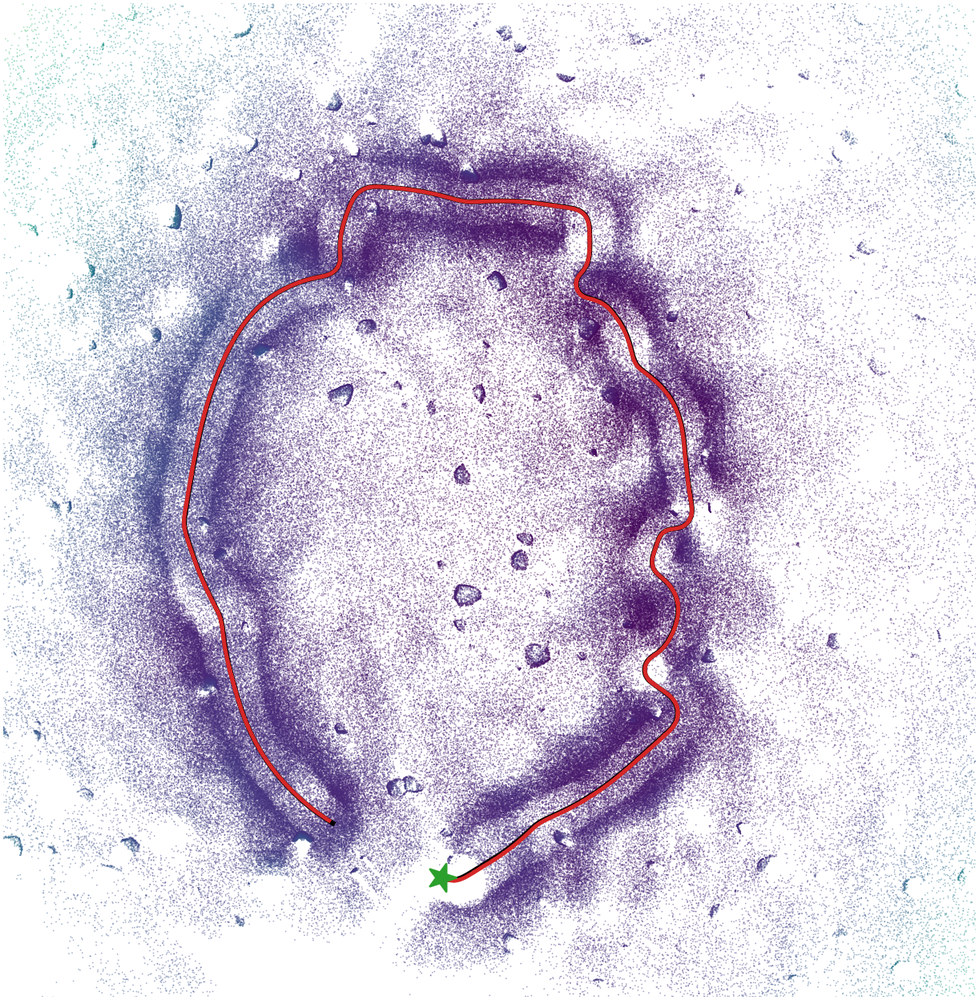} \\
    \slamrlab{Grand Canyon}
      & \slamfail & \slamfail
      & \slamtile{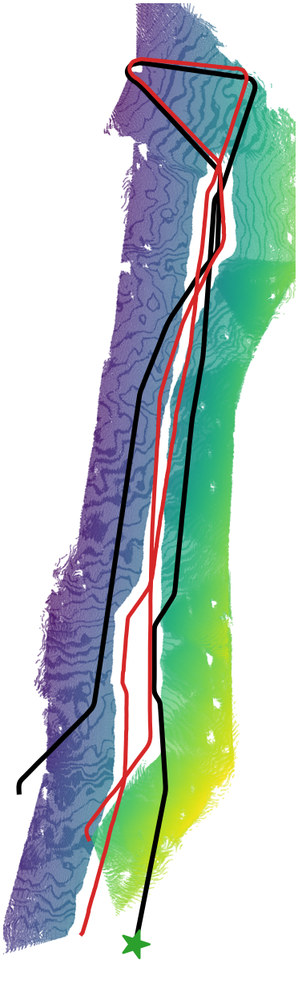} & \slamtile{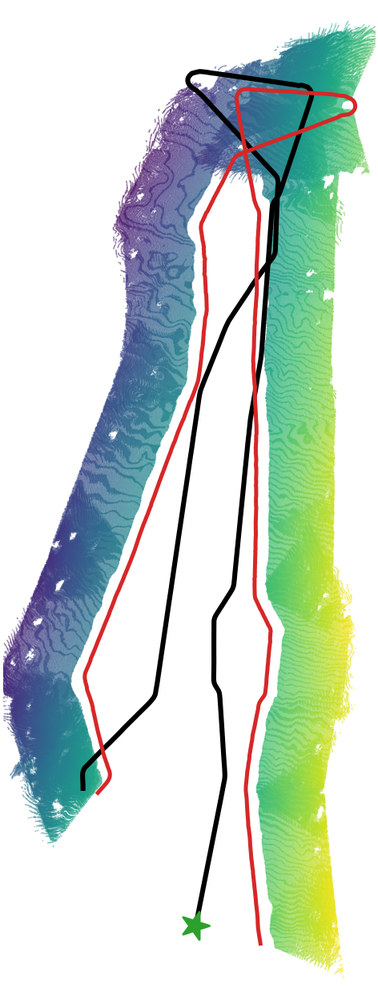}
      & \slamtile{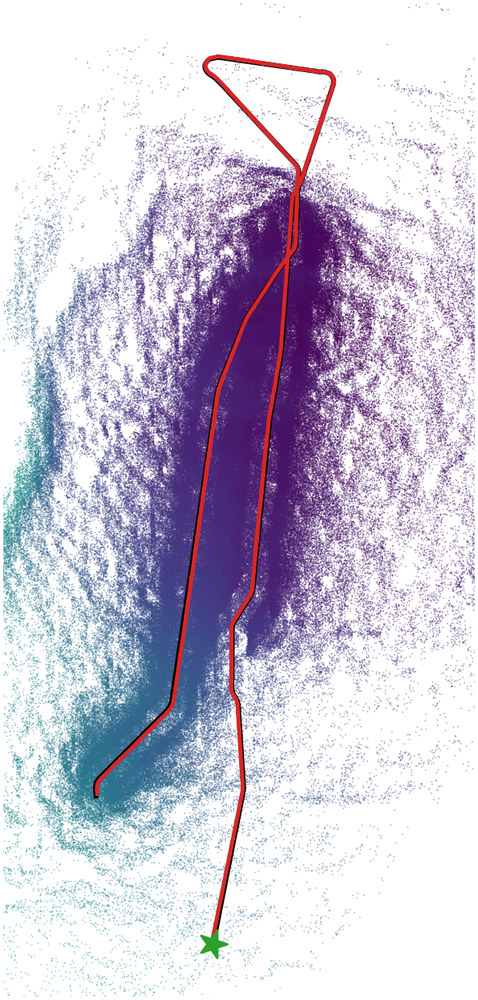} \\
  \end{tabular}
  \caption{\textbf{Multi-scene SLAM benchmark: estimated vs.\ ground-truth
  trajectories.} Top-down view of every run---the estimated trajectory (red)
  versus ground truth (black) after Umeyama alignment, over the reconstructed
  point-cloud map (coloured by height); rows are scenes, columns are sensing
  modality and dust optical depth. Grey tiles (\no) mark monocular ORB-SLAM runs
  that lost tracking and produced no usable trajectory. Scene names and GT
  trajectory lengths are listed in Table~\ref{tab:slam-gt-trajectories}; GT
  trajectory geometry is shown
  in \figref{fig:slam-gt}; per-run ATE in Table~\ref{tab:slam-ate};
  reconstructed 3D maps in \figref{fig:slam-maps}.}
  \label{fig:slam-traj}

  \vspace{0.9em}

  \setlength{\tabcolsep}{1.5pt}%
  \renewcommand{\arraystretch}{1.0}%
  \begin{tabular}{@{}c c c c@{}}
    & \textbf{Mars Base} & \textbf{Main Crater Interior} & \textbf{Grand Canyon} \\[1pt]
    \maprlab{Visual SLAM}
      & \maptile{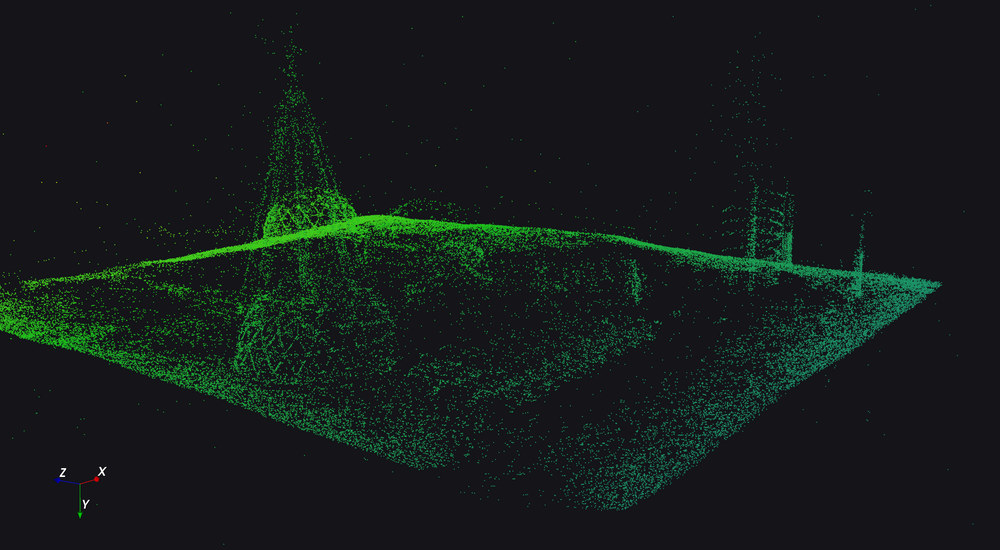} & \maptile{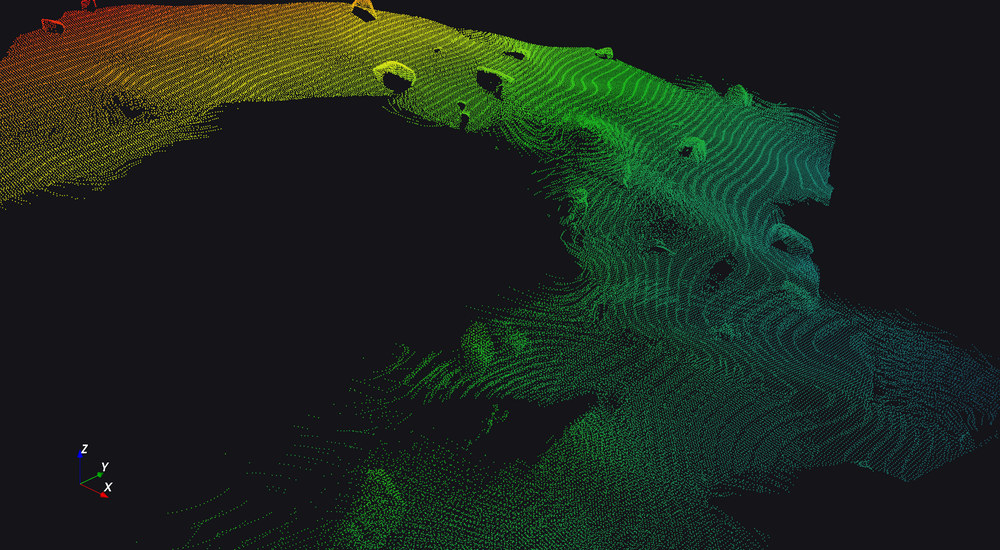} & \maptile{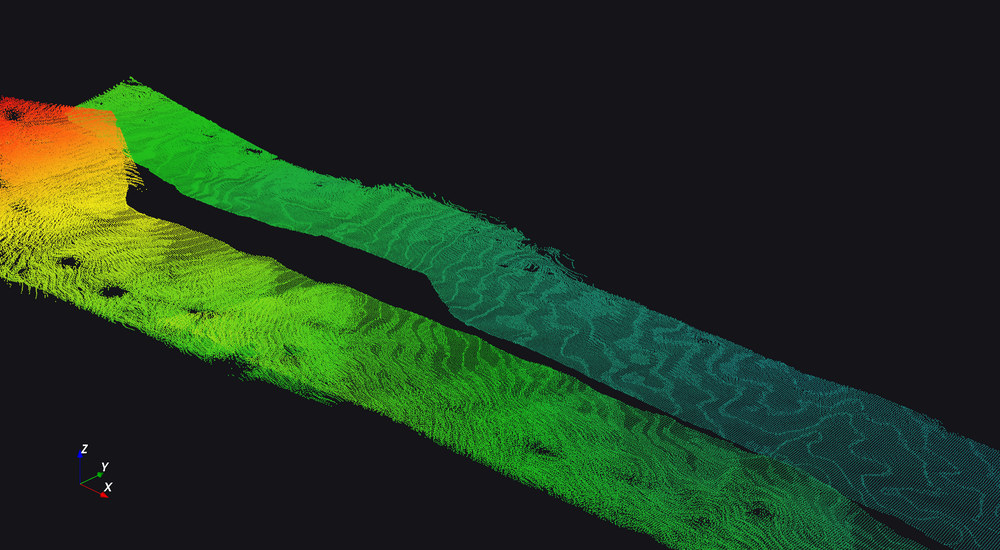} \\
    & {\scriptsize ORB-SLAM (RGB)} & {\scriptsize RTAB-Map (RGB-D+Odom)}
      & {\scriptsize RTAB-Map (RGB-D+Odom)} \\[2pt]
    \maprlab{LiDAR SLAM}
      & \maptile{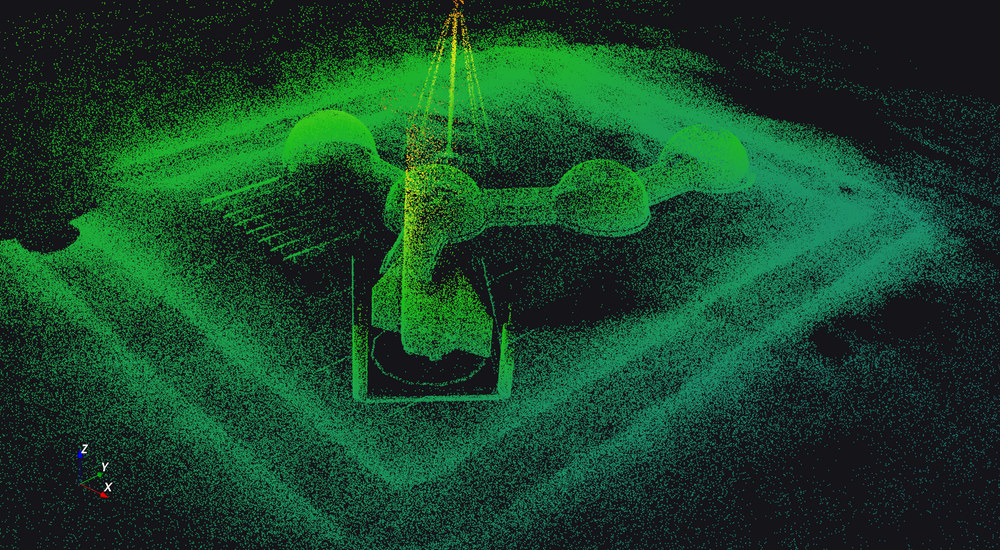} & \maptile{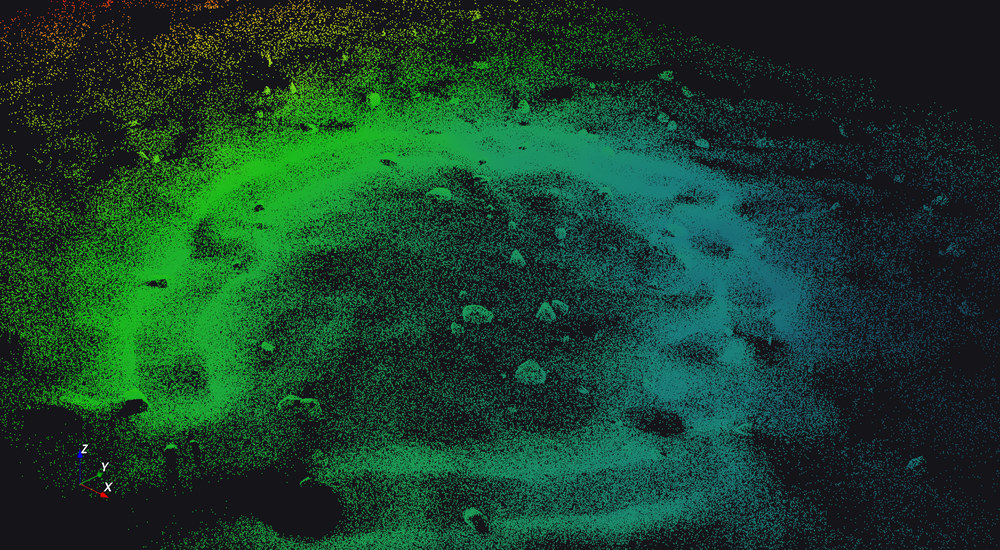} & \maptile{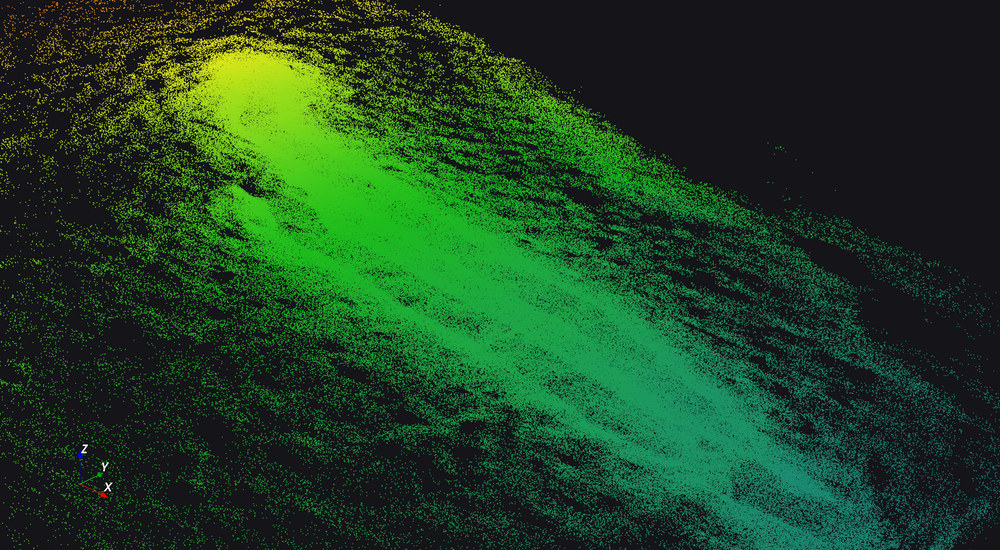} \\
    & {\scriptsize MOLA (LiDAR)} & {\scriptsize MOLA (LiDAR)}
      & {\scriptsize MOLA (LiDAR)} \\[1pt]
    & {\scriptsize (a)} & {\scriptsize (b)} & {\scriptsize (c)} \\
  \end{tabular}
  \caption{\textbf{Reconstructed 3D maps: Visual SLAM vs.\ LiDAR SLAM.}
  Point-cloud maps reconstructed for each scene, comparing visual SLAM (top)
  against LiDAR SLAM (bottom; MOLA). For (a) \emph{Mars Base} the visual-SLAM map is
  from ORB-SLAM (RGB); for (b) \emph{Main Crater Interior} and (c) \emph{Grand Canyon} it is
  from RTAB-Map (RGB-D+Odom), whose depth and wheel-odometry inputs remained
  stable where monocular ORB-SLAM lost tracking. Points are coloured by
  elevation.}
  \label{fig:slam-maps}
\tightfloatvspace
\end{figure*}

\section{Experiments}
\label{sec:experiments}

We evaluate MarsLab with two studies: a multi-scene \ac{SLAM} benchmark comparing
RGB, RGB-D/wheel-odometry, and 3D-LiDAR methods under controlled dust, and a
\ac{VPR} benchmark over repeated \emph{Mars Base} traversals.

\subsection{\texorpdfstring{\ac{SLAM}}{SLAM} Benchmark}
\label{sec:exp-slam}

\noindent\textbf{Setup.}  The \ac{SLAM} benchmark uses three representative scenes: \emph{Mars
Base} for flat terrain with structural landmarks, \emph{Main Crater Interior}
for rock-scattered crater terrain, and \emph{Grand Canyon} for large elevation
changes and occlusion.  We evaluate ORB-SLAM~\cite{campos2021orbslam3} on monocular RGB,
RTAB-Map~\cite{labbe2019rtabmap} on RGB-D plus wheel odometry, and
MOLA~\cite{blanco2019mola} on 3D LiDAR.  ORB-SLAM and RTAB-Map are run at
$\tau{=}0.5$ and $\tau{=}6.0$; MOLA is run once per scene because the dust model
affects radiance but not range.

\noindent\textbf{Protocol.}  For each scene, we design the \ac{GT} trajectory to keep
the rover on traversable terrain by accounting for rock distribution and local
terrain slope.  The pure-pursuit follower executes the trajectory, and MarsLab
logs ground truth poses in TUM format.  We report \ac{ATE}
RMSE after Umeyama SE(3) alignment.  Table~\ref{tab:slam-gt-trajectories} and
\figref{fig:slam-gt} define the \ac{GT} trajectories, \figref{fig:slam-traj} overlays
trajectories, and \figref{fig:slam-maps} visualizes reconstructed maps.  All
runs use Isaac~Sim~5.1, ROS2~Jazzy, and an NVIDIA~RTX~5070~Ti workstation.

\begin{table}[tbp]
  \centering
  \begin{threeparttable}
  \caption{ATE RMSE (m) across MarsLab scenes and SLAM stacks.  Camera
  methods are run at low ($\tau{=}0.5$) and high ($\tau{=}6.0$) dust; LiDAR is
  run once per scene.  Per-row best in \textbf{bold}; \no\ marks a tracking
  failure.}
  \label{tab:slam-ate}
  \scriptsize
  \setlength{\tabcolsep}{1.8pt}
  \renewcommand{\arraystretch}{1.02}
  \begin{tabular*}{\columnwidth}{@{\extracolsep{\fill}}lccccc@{}}
  \toprule
   & \multicolumn{2}{c}{ORB-SLAM} & \multicolumn{2}{c}{RTAB-Map} & MOLA \\
  \cmidrule(lr){2-3}\cmidrule(lr){4-5}\cmidrule(lr){6-6}
  Scene & $\tau{=}0.5$ & $\tau{=}6.0$ & $\tau{=}0.5$ & $\tau{=}6.0$ & LiDAR \\
  \midrule
  Mars Base & 0.59 & 4.70 & 0.48 & 0.64 & \textbf{0.20} \\
  \makecell[l]{Main Crater\\Interior} & 40.28 & \no & 8.84 & 10.61 & \textbf{0.13} \\
  Grand Canyon & \no & \no & 12.28 & 19.08 & \textbf{0.38} \\
  \bottomrule
  \end{tabular*}
  \begin{tablenotes}[flushleft]\scriptsize
  \item \no\ = monocular tracking failure (no usable trajectory).
  \end{tablenotes}
  \end{threeparttable}
\tightfloatvspace
\end{table}

%

%
\begin{figure}[tbp]
  \centering
  \newcommand{\gtw}{0.305\columnwidth}%
  \newcommand{\routeicon}{\tikz[baseline=-0.45ex]{\draw[black, line width=0.55pt] (0,0) -- (0.55,0);}}%
  \newcommand{\starticon}{\tikz[baseline=-0.70ex]{\node[star, star points=5, star point ratio=2.25, fill=green!75!black, draw=white, line width=0.25pt, minimum size=5.0pt, inner sep=0pt] {};}}%
  \setlength{\tabcolsep}{1.0pt}%
  \renewcommand{\arraystretch}{0.90}%
  \begin{tabular}{@{}ccc@{}}
    \includegraphics[width=\gtw]{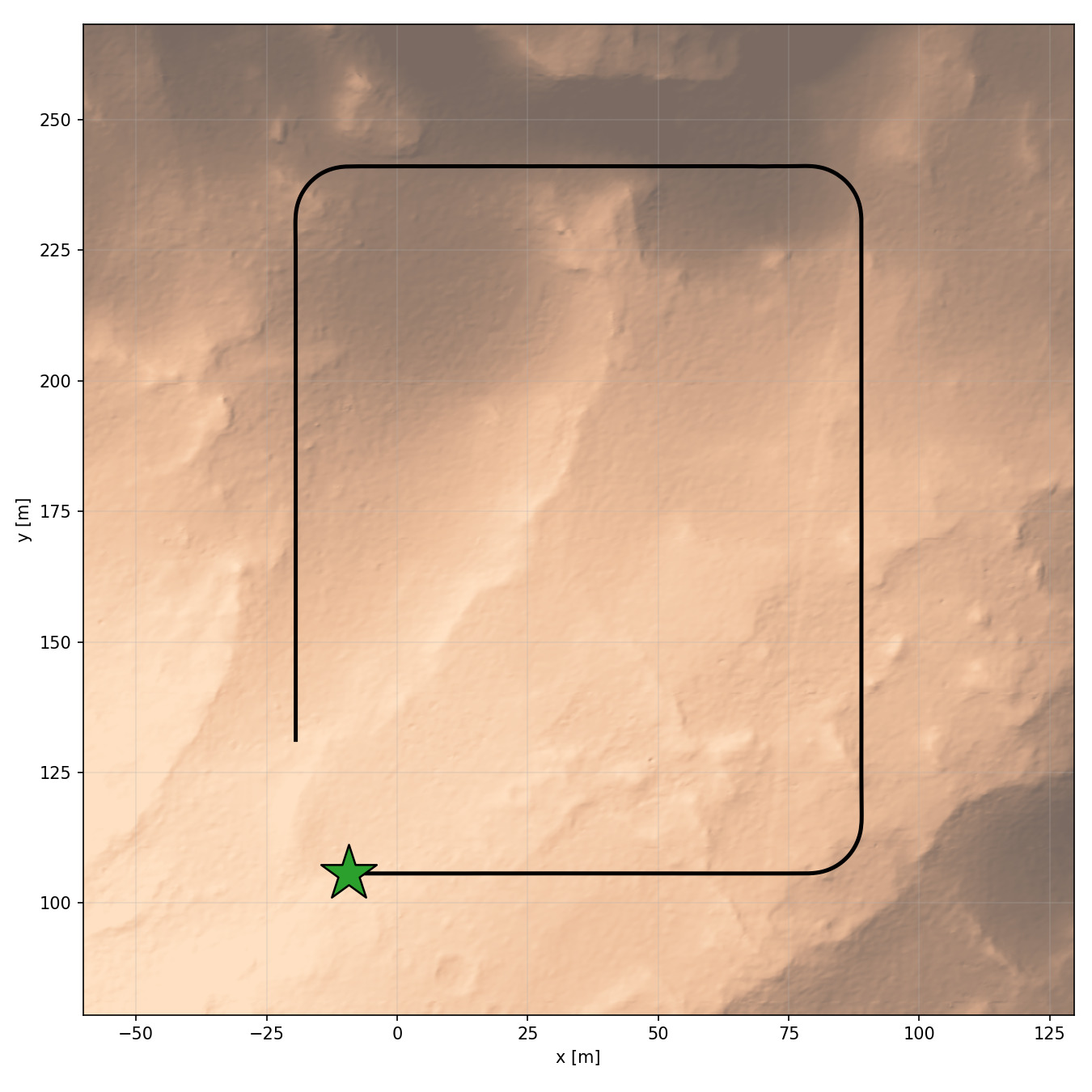} &
    \includegraphics[width=\gtw]{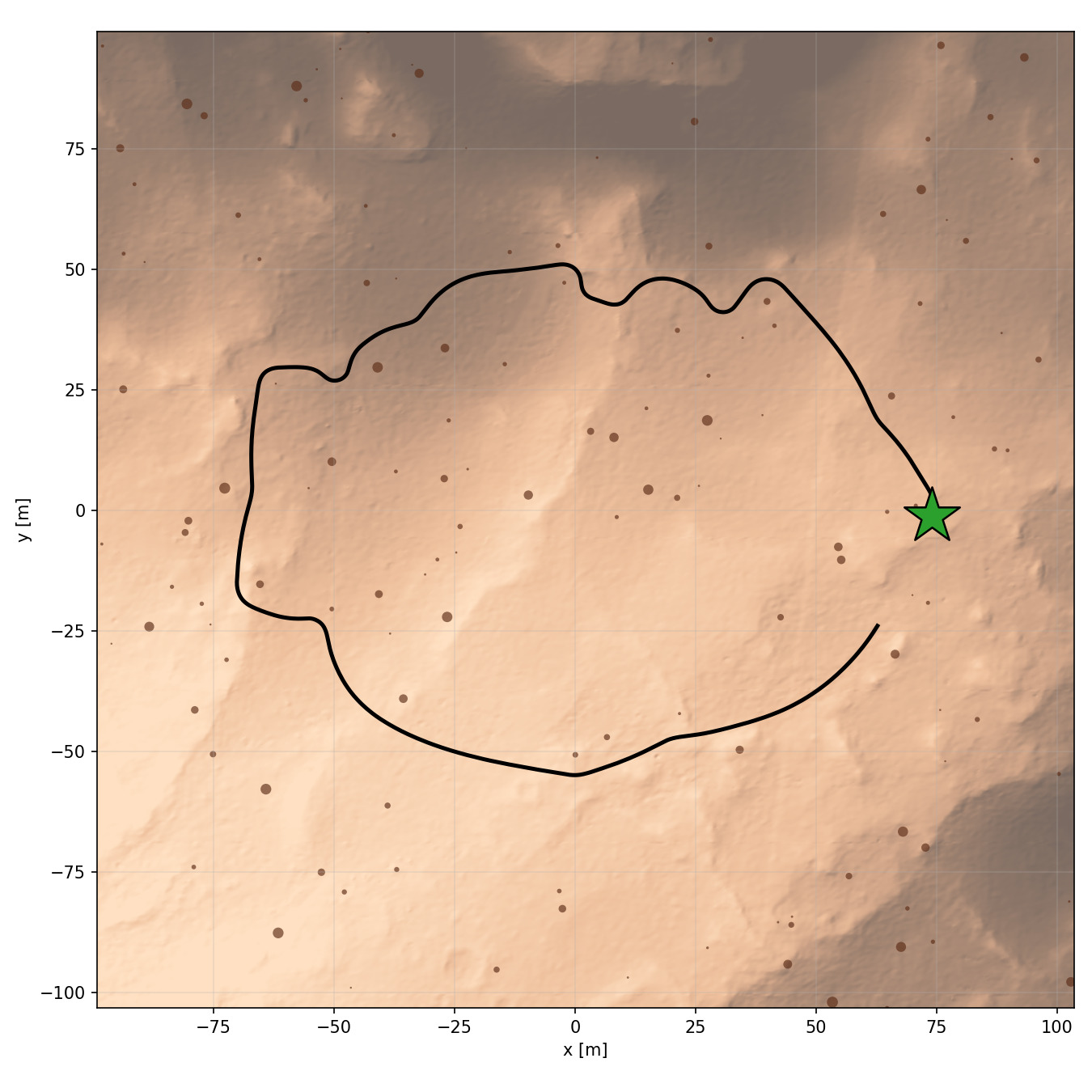} &
    \includegraphics[width=\gtw]{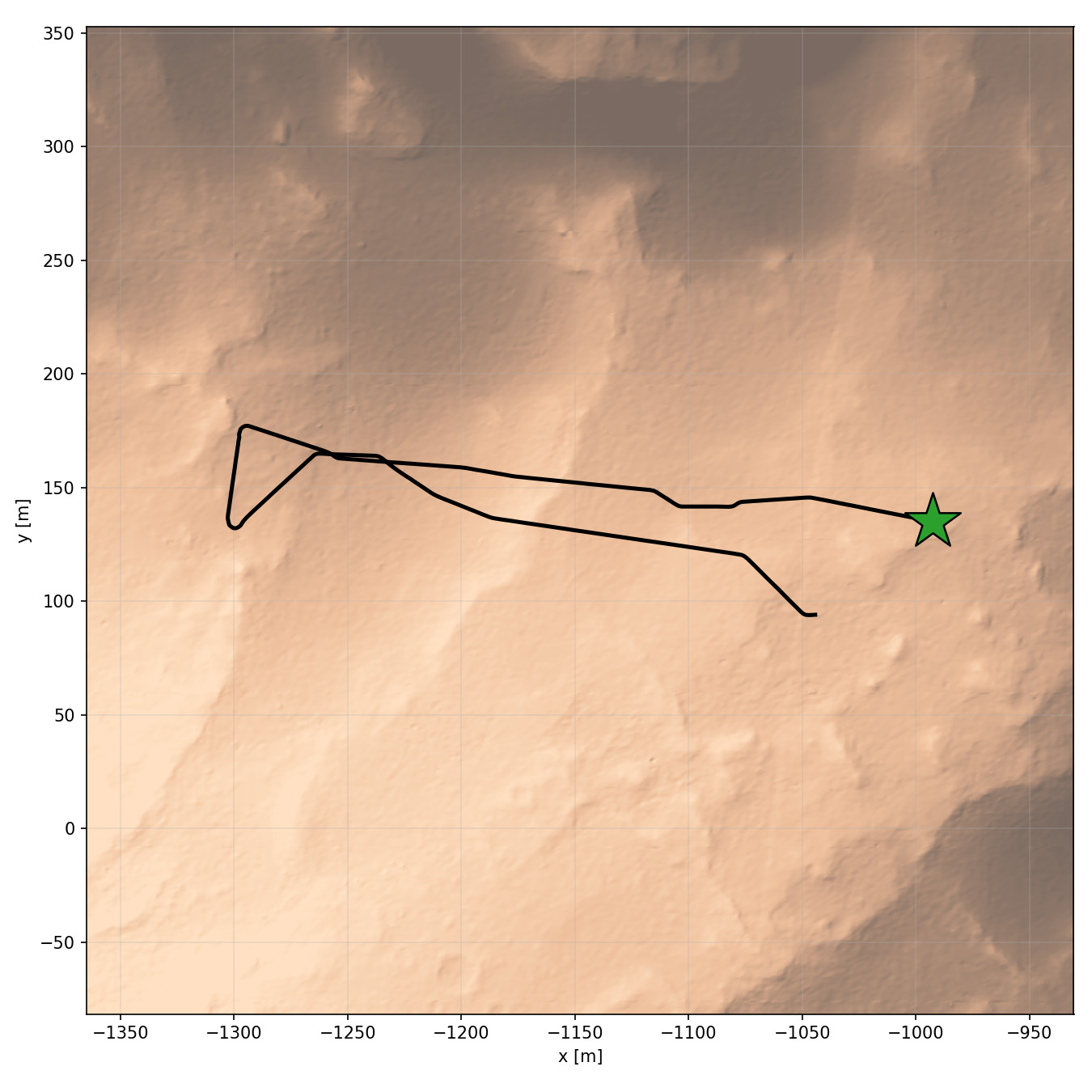} \\
    {\scriptsize Mars Base} & {\scriptsize Main Crater} & {\scriptsize Grand Canyon} \\
    \multicolumn{3}{c}{\scriptsize \routeicon\ route \quad \starticon\ start point}
  \end{tabular}
  \caption{\textbf{GT trajectories.} Compact overview of the three indexed
  ground-truth trajectories used in the SLAM benchmark.}
  \label{fig:slam-gt}
\tightfloatvspace
\end{figure}
\begin{figure}[tbp]
  \centering
  \captionof{table}{\textbf{SLAM benchmark GT trajectory metadata.}
  Each GT trajectory is authored in a MarsLab scene, replayed by the rover
  runtime, and logged from Isaac~Sim pose streams in TUM format for ATE/RPE
  evaluation.}
  \label{tab:slam-gt-trajectories}
  \scriptsize
  \setlength{\tabcolsep}{1.3pt}%
  \renewcommand{\arraystretch}{0.98}%
  \begin{tabularx}{\columnwidth}{@{}>{\raggedright\arraybackslash}p{0.25\columnwidth}@{\hspace{0.2em}}>{\centering\arraybackslash}p{0.13\columnwidth}@{\hspace{1.0em}}>{\raggedright\arraybackslash}X@{}}
    \toprule
    \textbf{Scene} & \textbf{Length} & \mbox{\textbf{GT Trajectory Role}} \\
    \midrule
    Mars Base & $\approx440$\,m & Feature- and landmark-rich trajectory for testing localization in structured habitat scenes. \\
    \makecell[l]{Main Crater\\Interior} & $\approx393$\,m & Rock-rich trajectory for testing localization under repetitive terrain and sparse visual features. \\
    \addlinespace[0.70em]
    Grand Canyon & $\approx648$\,m & Long-distance trajectory for testing drift across extended canyon terrain with large elevation changes. \\
    \bottomrule
  \end{tabularx}
\tightfloatvspace
\end{figure}

\noindent\textbf{Results.}  Table~\ref{tab:slam-ate} shows a clear modality ordering.
MOLA remains sub-metre on all scenes ($0.13$--$0.38$\,m), RTAB-Map ranges from
$0.48$\,m on \emph{Mars Base} to $8$--$19$\,m on rugged scenes, and monocular ORB-SLAM
is fragile, reaching $40.3$\,m on \emph{Main Crater Interior} and losing tracking on
\emph{Grand Canyon}.  Increasing dust from $\tau{=}0.5$ to $\tau{=}6.0$ mainly affects
vision: ORB-SLAM grows $7.9\times$ on \emph{Mars Base} and fails on \emph{Main Crater
Interior}, while RTAB-Map degrades more gradually because depth and wheel odometry
anchor scale and motion.  MOLA is unchanged, consistent with a dust model that
attenuates radiance but not range.

\noindent\textbf{Failure modes.}  Scene geometry also matters.  \emph{Mars Base} is easiest for
vision, whereas canyon walls and self-similar crater terrain starve monocular
features, causing the tracking losses marked in \figref{fig:slam-traj} and
Table~\ref{tab:slam-ate}.  MOLA's error instead follows \ac{GT} trajectory length,
rising from $0.13$\,m on the $393$\,m \emph{Main Crater Interior} trajectory to
$0.38$\,m on the $648$\,m \emph{Grand Canyon} trajectory, or only $0.03$--$0.06\%$ of
distance travelled.
This separation of dust, geometry, and \ac{GT} trajectory length illustrates the
controlled attribution MarsLab is designed to support.

\subsection{Visual Place Recognition}
\label{sec:exp-vpr}

\noindent\textbf{Setup.}  Beyond trajectory-level \ac{SLAM}, we evaluate \ac{VPR} as
image retrieval over repeated \emph{Mars Base} traversals.  We compare
NetVLAD~\cite{netvlad}, AnyLoc~\cite{anyloc}, and BoQ~\cite{boq} on $4456$
query and $4455$ database frames recorded every $0.1$\,m.  A retrieval is
correct when one of the top-$N$ database images lies within $2$\,m of the query
pose; we report Recall@$N$ for $N\in\{1,5,10\}$.

\noindent\textbf{Protocol.}  We test two appearance changes on the same Mars
Base route.  \emph{Day/Dark} matches day-lit queries to a dark database at
$\tau{=}0.05$, while \emph{Dust} matches low-dust daytime queries at
$\tau{=}0.05$ to high-dust daytime database images at $\tau{=}0.60$.

\begin{figure}[tbp]
  \centering
  \setlength{\tabcolsep}{1.5pt}%
  \renewcommand{\arraystretch}{1.0}%
  \newcommand{\vprw}{0.235\columnwidth}%
  \begin{tabular}{@{}cccc@{}}
    {\scriptsize Query} & {\scriptsize Retrieved (R@1)} &
    {\scriptsize Query} & {\scriptsize Retrieved (R@1)} \\
    \includegraphics[width=\vprw]{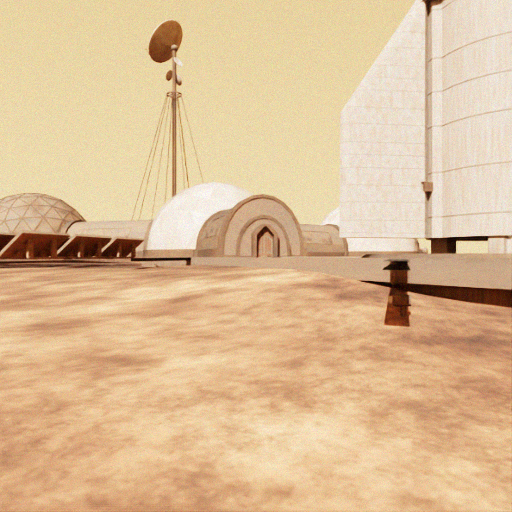} &
    \includegraphics[width=\vprw]{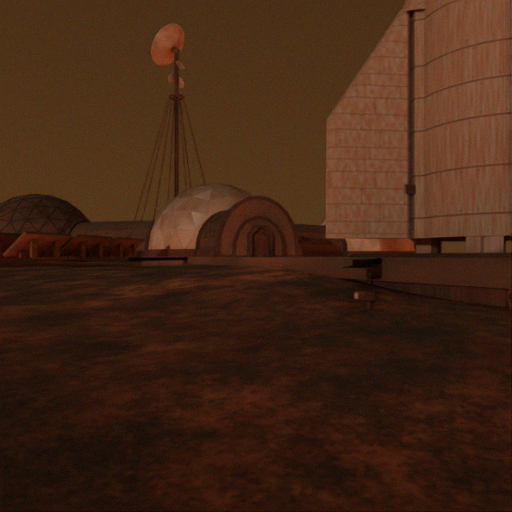} &
    \includegraphics[width=\vprw]{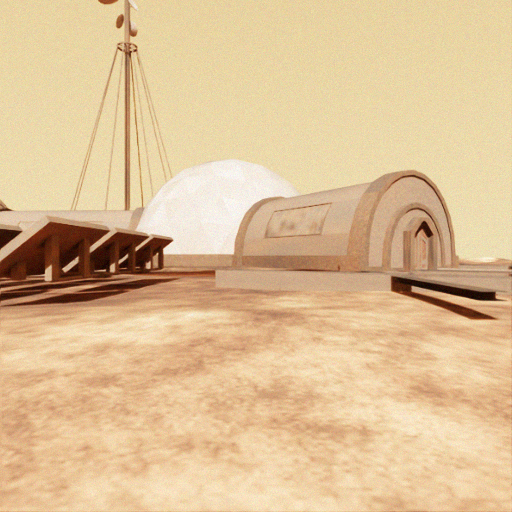} &
    \includegraphics[width=\vprw]{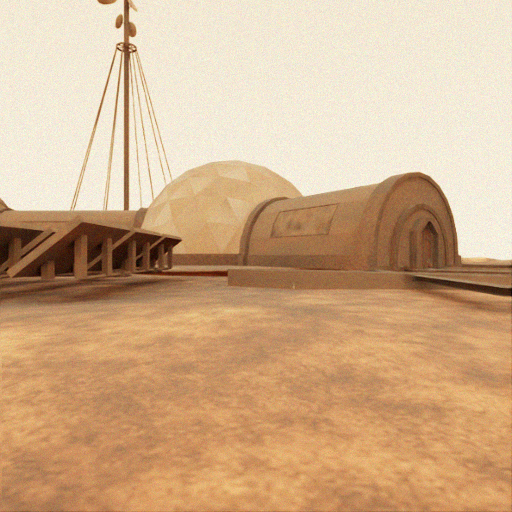} \\
    \multicolumn{2}{c}{\footnotesize (i) Day/Dark, $\tau{=}0.05$} &
    \multicolumn{2}{c}{\footnotesize (ii) Dust, $\tau{=}0.05$ to $\tau{=}0.60$} \\
  \end{tabular}
  \caption{\textbf{Visual place recognition: query vs. retrieved.}  Qualitative
  examples for the two Table~\ref{tab:vpr-recall} protocols: (i) day-lit queries
  matched to a dark database at $\tau{=}0.05$ and (ii) low-dust daytime queries
  at $\tau{=}0.05$ matched to high-dust daytime database images at
  $\tau{=}0.60$.  Each pair shows the query frame and the top-1 retrieved
  database frame (R@1).}
  \label{fig:vpr}
\tightfloatvspace
\end{figure}

\noindent\textbf{Results.}  Table~\ref{tab:vpr-recall} shows that BoQ performs
best in both protocols, reaching $91.44\%$ R@1 for Day/Dark and $94.65\%$ R@1
for Dust.  NetVLAD ranks second across all metrics, and AnyLoc recovers by R@10
despite lower R@1 accuracy.  \figref{fig:vpr} confirms the same trend
qualitatively: the top-1 matches for the habitat dome and lander/Starship still
depict the correct places under viewpoint and appearance changes.

\begin{table}[tbp]
\centering
\caption{Visual place recognition recall on the MarsLab Mars-base benchmark
using the strict $2$\,m positive radius. Results are Recall@$N$ (\%). Best
values in each protocol/metric are shown in bold.}
\label{tab:vpr-recall}
\scriptsize
\setlength{\tabcolsep}{2.0pt}
\renewcommand{\arraystretch}{1.10}
\begin{tabular*}{\columnwidth}{@{\extracolsep{\fill}}lcccccc@{}}
\toprule
\multirow{2}{*}{Method}
& \multicolumn{3}{c}{Day/Dark, $\tau{=}0.05$}
& \multicolumn{3}{c}{Dust, $\tau{=}0.05{\rightarrow}0.60$} \\
\cmidrule(lr){2-4}\cmidrule(lr){5-7}
& R@1 & R@5 & R@10 & R@1 & R@5 & R@10 \\
\midrule
NetVLAD~\cite{netvlad}
& 77.90 & 95.35 & 98.36
& 85.02 & 98.18 & 99.33 \\
AnyLoc~\cite{anyloc}
& 64.04 & 92.21 & 97.17
& 74.89 & 95.71 & 99.06 \\
BoQ~\cite{boq}
& \textbf{91.44} & \textbf{99.19} & \textbf{99.80}
& \textbf{94.65} & \textbf{99.37} & \textbf{99.78} \\
\bottomrule
\end{tabular*}
\tightfloatvspace
\end{table}

\subsection{Experiment Summary}
\label{sec:exp-summary}

Taken together, the experiments show how MarsLab supports autonomy benchmarks
across both trajectory-level estimation and image-level place recognition.  The
\ac{SLAM} benchmark varies dust level, long-range \ac{GT} trajectory, rock distribution,
and scene structure, spanning feature- and landmark-poor terrain as well as
landmark-rich \emph{Mars Base} conditions.  It also exposes the sensing modalities
needed for practical rover navigation: monocular camera, RGB-D, LiDAR, and wheel
odometry, with IMU streams available for additional tests even though this paper
does not include an IMU-based \ac{SLAM} run.  The \ac{VPR} benchmark complements this by
using the same \emph{Mars Base} route to change dust and illumination while preserving
pose labels, allowing users to design controlled \ac{VPR} protocols and compare
retrieval behavior across methods.  The different R@1 scores in
Table~\ref{tab:vpr-recall} confirm that these appearance changes produce
meaningful algorithm-dependent differences rather than trivial retrieval cases.

\section{Discussion}
\label{sec:discussion}

MarsLab combines customizable Mars environments with ROS2 sensor interfaces
and ground-truth outputs, enabling repeatable autonomy experiments under
different terrain, illumination, and dust settings.  Future work will extend
the IMU noise model to include constant biases and time-varying drift, and
incorporate particle-based terramechanics for studying deformable regolith
and wheel--soil interaction.

In addition to these modeling extensions, future work will introduce multi-robot and heterogeneous platform support, including
different rover types and aerial scouts with platform-specific dynamics and
controllers.  These extensions will broaden evaluation beyond \ac{SLAM} and
\ac{VPR} to computer-vision tasks, traversability estimation, path planning,
and multi-robot cooperation, using shared Mars environments, customizable
sensor interfaces, and per-robot ground-truth outputs.

\section{Conclusion}
\label{sec:conclusion}

We presented MarsLab, an open-source, ROS2-native Mars rover simulator for
reproducible autonomy evaluation.  MarsLab combines HiRISE-derived and
procedural terrain with distributed rock, crater, illumination, and dust models,
then plays each scene and \ac{GT} trajectory through an Isaac~Sim/ROS2 runtime with
ground truth pose data.  The experiments show that MarsLab lets users vary sensing
modality, dust, scene geometry, and \ac{GT} trajectory length while keeping the
remaining evaluation conditions explicit.  By coupling Mars-relevant scene
generation with rover motion and standard robotics interfaces, MarsLab provides
a practical testbed for localization, mapping, navigation, place recognition,
and broader rover-autonomy algorithm development.

\bibliographystyle{IEEEtran}
\bibliography{ref}

\end{document}